\documentclass[lettersize,journal]{IEEEtran}
\usepackage{textcomp}
\usepackage{stfloats}
\usepackage{url}
\usepackage{verbatim}
\usepackage{graphicx}
\usepackage[nocompress]{cite}
\usepackage{caption}
\usepackage{subcaption}
\usepackage{amsmath}
\usepackage{amsfonts}
\usepackage{amssymb}
\usepackage{cancel}
\usepackage{algorithm}
\usepackage{algpseudocode}
\usepackage{multirow}
\usepackage{multicol}
\usepackage{adjustbox}
\usepackage{booktabs}
\usepackage{hhline}
\usepackage{bm}
\usepackage{adjustbox}
\usepackage{color}
\usepackage[colorlinks,
            linkcolor=blue,
            anchorcolor=blue,
            citecolor=blue]{hyperref}
\usepackage{indentfirst}
\usepackage{diagbox}
\usepackage{tabu}
\usepackage{makecell}
\usepackage[table,xcdraw]{xcolor}
\usepackage{ragged2e}
\usepackage{array}
\usepackage{tabularx}
\newcolumntype{L}[1]{>{\RaggedRight\arraybackslash}p{#1}}
\newcolumntype{C}[1]{>{\Centering\arraybackslash}p{#1}}
\newcolumntype{Y}{>{\Centering\arraybackslash}X}

\usepackage[T1]{fontenc}
\definecolor{mygray}{gray}{.9}
\usepackage{pifont}

\usepackage{xspace}
\makeatletter
\DeclareRobustCommand\onedot{\futurelet\@let@token\@onedot}
\def\@onedot{\ifx\@let@token.\else.\null\fi\xspace}

\def\eg{\emph{e.g}\onedot} 
\def\ie{\emph{i.e}\onedot}

\makeatother

\usepackage[table]{xcolor}

\definecolor{topOneGreen}{HTML}{78B97A}
\definecolor{topTwoGreen}{HTML}{DDEEDD}

\newcommand{\topone}[1]{\cellcolor{green!45}\textbf{#1}}
\newcommand{\toptwo}[1]{\cellcolor{green!20}#1}

\usepackage{pifont}
\definecolor{softgreen}{RGB}{46,125,50}
\definecolor{softred}{RGB}{198,40,40}
\newcommand{\cmark}{\textcolor{softgreen}{\ding{51}}}
\newcommand{\xmark}{\textcolor{softred}{\ding{55}}}

\usepackage{enumitem}
\usepackage{multirow}
\makeatletter
\newcommand{\thickhline}{%
 \noalign {\ifnum 0=`}\fi \hrule height 1pt
 \futurelet \reserved@a \@xhline
}
\makeatother

\usepackage{amsmath,amsfonts,bm}

\def\eqref#1{equation~\ref{#1}}
\def\1{\bm{1}}

\def\mM{{\bm{M}}}

\def\mO{{\bm{O}}}

\def\mX{{\bm{X}}}
\def\mY{{\bm{Y}}}

\DeclareMathAlphabet{\mathsfit}{\encodingdefault}{\sfdefault}{m}{sl}
\SetMathAlphabet{\mathsfit}{bold}{\encodingdefault}{\sfdefault}{bx}{n}

\def\sR{{\mathbb{R}}}

\begin{document}
\bstctlcite{BSTcontrol}

\title{
% Beyond Shadow Masks: Subject-aware Reasoning for Controllable Shadow Removal
%FreeDeshadow: Prompt-aware Reasoning for Controllable Shadow Removal
FreeDeshadow: Shadow Removal for User-Specified Objects With Multimodal Prompts
}

\author{
Zhiliang Wu,~\IEEEmembership{Member,~IEEE,}
Jihao Gu, 
Kun Li,~\IEEEmembership{Member,~IEEE,}
Hanyu Xuan,
Hehe Fan,~\IEEEmembership{Senior Member,~IEEE,}\\
Meng Wang,~\IEEEmembership{Fellow,~IEEE,}
Weisi Lin, \emph{Fellow, ~IEEE}
        % <-this % stops a space
%\thanks{This paper was produced by the IEEE Publication Technology Group. They are in Piscataway, NJ.}% <-this % stops a space
\thanks{Manuscript received November 5, 2026.  %This work was supported in part by the National Natural Science Foundation of China (92570101).
}
\thanks{(Corresponding author: \textit{Kun Li}.)}
\IEEEcompsocitemizethanks
{
\IEEEcompsocthanksitem 
Zhiliang Wu and Weisi Lin are with the College of Computing and Data Science, Nanyang Technological University, Singapore 639798, Singapore.% (e-mail: zhiliang.wu@ntu.edu.sg and wslin@ntu.edu.sg).
\IEEEcompsocthanksitem
Jihao Gu is with the Department of Computer Science, University College London, London WC1E 6BT, U.K.% (e-mail: jihao.gu.23@ucl.ac.uk).
\IEEEcompsocthanksitem
Kun Li is with the College of Computing and Artificial Intelligence, United Arab Emirates University, Abu Dhabi 15551, United Arab Emirates.% (e-mail: kun.li@uaeu.ac.ae).
\IEEEcompsocthanksitem
Hanyu Xuan is with the School of Big Data and Statistics, Anhui University, Hefei 230039, China.% (e-mail: 22176@ahu.edu.cn).
\IEEEcompsocthanksitem
Hehe Fan is with the School of Artificial Intelligence, Zhejiang University, Hangzhou 310007, China.% (e-mail: hehefan@zju.edu.cn).
\IEEEcompsocthanksitem
Meng Wang is with the School of Computer Science and Information Engineering, Hefei University of Technology, Hefei 230601, China.% (e-mail: eric.mengwang@gmail.com).
%\IEEEcompsocthanksitem * Corresponding author.
}
}
% The paper headers
\markboth{IEEE Transactions on xxx,~Vol.~, No.~, October~2026}%
{Shell \MakeLowercase{\textit{et al.}}: A Sample Article Using IEEEtran.cls for IEEE Journals}

\maketitle

\begin{abstract}
Existing shadow removal methods either perform global shadow removal or rely on precise shadow masks to guide restoration. 
Such paradigms limit their ability to selectively remove the shadow cast by a user-specified object in practical applications. 
To this end, we introduce an object-aware controllable shadow removal task, which allows users to specify a target object through a simple prompt and removes only its cast shadow while preserving other content. 
However, object–shadow asymmetry makes it challenging to establish accurate correspondence between the specified object and its spatially separated cast shadow and to confine restoration to the target shadow region.
To address these challenges, we propose FreeDeshadow, a unified controllable shadow removal framework that supports both visual and textual prompts. 
It consists of an object-aware generator and a shadow-aware local refiner. 
The former interprets the user prompt and produces a high-quality shadow-free proposal, while the latter establishes the correspondence between the object and its cast shadow and constrains the generated changes to the target shadow region.
To support training and evaluation, we further construct PCSRD+, a large-scale dataset containing 154,571 quadruples at 1024$\times$1024 resolution and covering six types of user prompts. 
The visual prompts include a dot, a line, and an object mask, while the textual prompts describe the target object using attributes, spatial relations, or states.
Extensive experiments demonstrate that FreeDeshadow accurately removes the shadow of the target object under both visual and textual prompts while effectively preserving non-target regions.
It also outperforms existing methods in shadow removal performance and controllability.
\textit{Project Page}: \href{https://wzlwlm13.github.io/FreeDeshadow/}{https://FreeDeshadow.github.io}.
\end{abstract}


\section{Introduction}
\IEEEPARstart{S}{hadows} are ubiquitous in natural images, arising when an object partially or fully occludes a light source~\cite{le2022physics}.
Although they provide useful geometric and illumination cues, cast shadows can undesirably obscure object appearances, distort surface colors and textures, and introduce misleading visual patterns. 
Such undesired shadows can degrade visual quality and hinder a wide range of applications, such as commercial product presentation, virtual staging, image editing, and content creation. 
These adverse effects have motivated extensive research on shadow removal~\cite{finlayson2009entropy,le2019shadowdecomposition}.
The task aims to recover the shadow-free appearance of a scene while faithfully preserving its underlying content. 
As a fundamental task in computational photography and computer vision, shadow removal~\cite{wang2018stcgan,wang2025softshadow,yu2025portrait,liu2023shadow,niu2023boundary,mei2024latent,wang2025metashadow} has remained an active research topic for decades.

\begin{figure}[!t]
\centering
\includegraphics[width=0.95\linewidth]{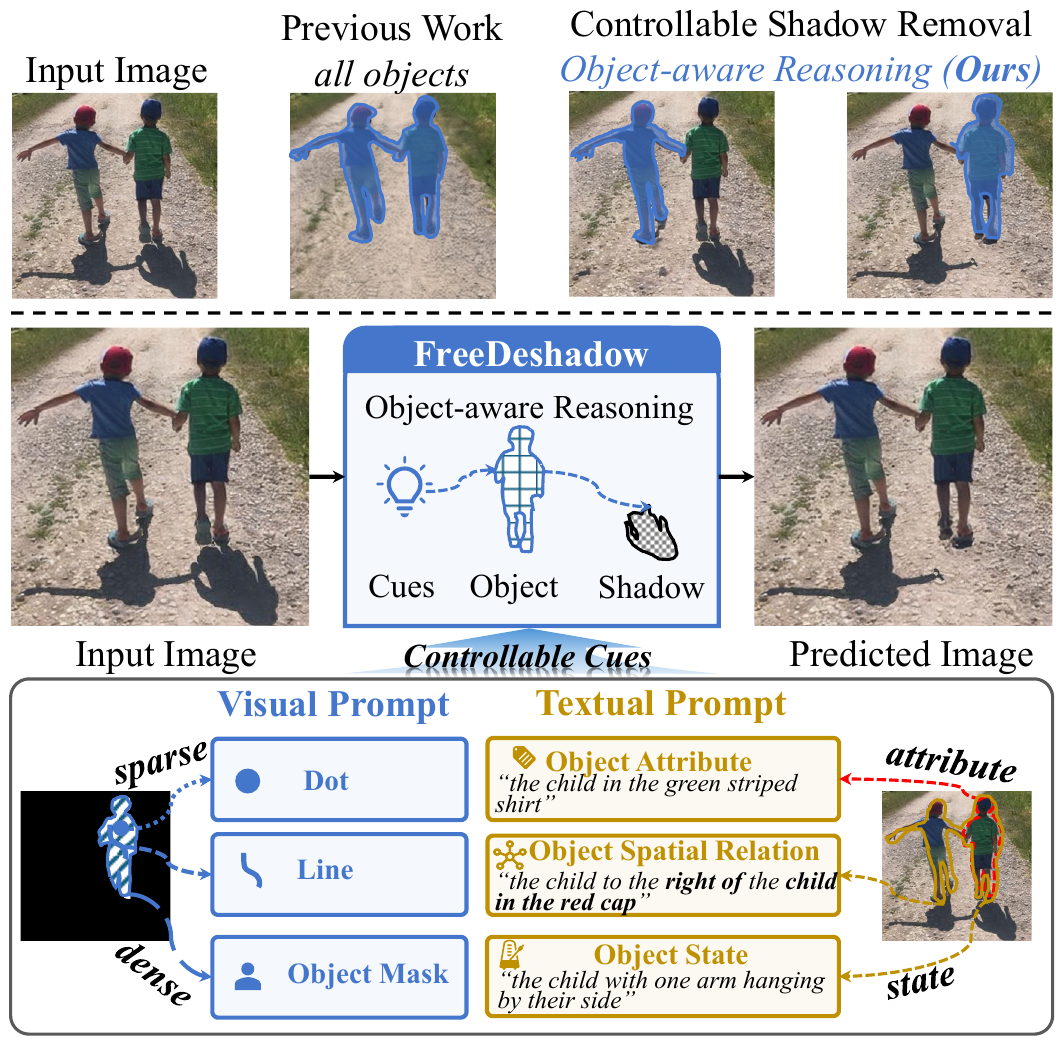}
\caption{Overview of object-aware controllable shadow removal. 
The object-aware shadow removal framework identifies the target object and selectively removes its cast shadow while preserving shadows belonging to other objects. Rather than relying on precise shadow-region masks as control inputs, FreeDeshadow enables flexible control through visual prompts of varying granularity (dots, lines, or object masks) and textual prompts describing attributes, spatial relations, or states.
}
\label{fig:teaser}
\end{figure}

Early shadow removal methods~\cite{finlayson2009entropy,finlayson2006removal,arbel2010shadow,yang2012shadow} largely rely on physical illumination models and hand-crafted priors.
They often fail to generalize to complex real-world scenes.
With the advent of deep learning, several deep shadow removal methods~\cite{wang2018stcgan,yu2025portrait,liu2023shadow,niu2023boundary,mei2024latent,wang2025metashadow} have been proposed and have achieved significant progress.
These methods can be divided into two paradigms: \emph{shadow-unaware} methods and \emph{shadow-aware} methods. 
The former~\cite{wang2018stcgan,yu2025portrait,liu2023shadow} take only the shadow image as input and remove shadows globally in an end-to-end manner. 
However, lacking reliable prior information on shadow locations, such methods struggle to distinguish shadow regions from inherently dark surfaces, leading to performance degradation in complex scenes.
The latter~\cite{niu2023boundary,mei2024latent,guo2023shadowformer} additionally introduce a precise shadow mask as input, using it as a pixel-level indicator of the shadow region to restore this region while preserving other content.
Conditioned on this prior, recent shadow-aware networks further leverage global context modeling~\cite{guo2023shadowformer,liu2024regional}, transformer~\cite{xiao2024homoformer,Polarization}, or diffusion priors~\cite{xu2025detail,luo2025diff} to improve restoration quality. 
Despite these advances, shadow-aware methods still suffer from two limitations. 
\textbf{First}, they rely heavily on high-quality annotated shadow masks. 
Manually annotating precise shadow regions is non-trivial because the boundaries between shadow and non-shadow regions are blurry. 
This process is also time-consuming and labor-intensive~\cite{chen2025promptaware}.
\textbf{Second}, they lack fine-grained control at the object level shadow removal.
In practice, users often prefer to specify an object directly and remove the shadow it casts.
For example, in a scene containing multiple objects, a user may wish to remove the shadow of one specific object, while preserving the object itself, the shadows of other objects, and the background.
However, the shadow mask only indicates where shadows are located, without revealing which object casts the shadow.
As a result, these methods cannot selectively remove the shadow cast by a user-specified object.

In this paper, we propose a new paradigm for shadow removal, termed \emph{\textbf{object-aware controllable shadow removal}}. 
Compared with shadow-aware methods, this paradigm allows users to directly specify the object of interest with a visual or textual prompt.
As shown in Fig.~\ref{fig:teaser}, given such a prompt, the model must infer the shadow cast by the specified object and remove only that shadow. 
Meanwhile, it should faithfully preserve the object itself, the shadows of other objects, and the background texture.
By shifting user interaction from shadow regions to the objects that cast them, this paradigm eliminates the need for users to provide precise shadow masks and enables more flexible object-level control.
Unlike conventional image editing~\cite{manukyan2025hdpainter,li2026brushedit,stabilityai2022sd15,zhang2025icedit,brooks2023instructpix2pix,fireredteam2025imageedit10}, object-aware controllable shadow removal involves an inherent discrepancy between the user-specified object and the region to be edited.
We refer to this property as \emph{object-shadow asymmetry}.
It introduces two challenges:
(1) an object and its cast shadow often occupy different spatial regions and differ substantially in appearance and shape. Consequently, the model must infer the shadow's location from geometric and illumination cues rather than directly treating the specified object as the editing region.
(2) shadows cast by different objects may share similar colors and intensities, and they may even overlap spatially.
Local visual features alone are insufficient to distinguish the target shadow from other shadows.
Therefore, the model must establish accurate object–shadow correspondence and confine the restoration to the target shadow region, thereby avoiding unintended modifications to other content.
More broadly, the spatial and appearance discrepancies underlying object–shadow asymmetry can also arise in other object-associated visual effects, such as reflections, caustics, and localized illumination. 
These effects pose a common challenge: the user-specified object and the region requiring modification may not coincide, necessitating accurate object–effect correspondence rather than direct prompt-to-region alignment. 
From this perspective, object-aware controllable shadow removal provides a concrete instance for investigating this broader challenge in controllable image editing.

Recent diffusion and flow-matching editing models~\cite{liu2025step1xedit,zhang2025icedit,paul2025fluxedit,wu2025qwenimage,mao2025aceplusplus,li2026brushedit} have demonstrated strong capabilities in semantic understanding and image editing, offering a promising direction for this task. 
However, directly applying these models to object-aware controllable shadow removal faces two limitations. 
\textit{On the one hand}, these models typically rely on image–prompt semantic alignment to identify editing targets. 
Due to object–shadow asymmetry, they tend to modify the specified object itself rather than infer and remove its cast shadow, resulting in unintended changes to the object's appearance.
\textit{On the other hand}, these models often perform editing through global image re-synthesis, which may alter the textures and colors of non-target regions even when only a localized shadow needs to be removed. Therefore, effective object-aware controllable shadow removal requires both accurate object–shadow correspondence and spatially constrained restoration to preserve non-target content.

To address these challenges, we propose FreeDeshadow, a unified object-aware controllable shadow removal framework that supports both visual and textual prompts. 
As illustrated in Fig.~\ref{fig:framework}, FreeDeshadow consists of an object-aware generator and a shadow-aware local refiner. 
The former leverages multimodal understanding and pretrained generative priors to recover the content obscured by shadows, while the latter addresses object–shadow asymmetry by establishing object–shadow correspondence and restricting restoration to the target shadow region. 
Specifically, the object-aware generator first interprets visual or textual prompts to derive a unified semantic representation of the specified object, capturing its identity and relevant contextual cues.
Conditioned on this representation, the generator exploits the powerful generative priors of a pretrained model to infer the textures, colors, and illumination obscured by the target shadow, producing a high-quality shadow-free proposal.
However, this proposal is generated through a global diffusion process, introducing unintended changes outside the target shadow region.
To address this limitation, the shadow-aware local refiner transforms the multimodal object representation into sparse shadow prompts, and employs cross-scale feature interactions to establish the correspondence between the target object and its spatially separated cast shadow. 
Guided by this correspondence, the refiner performs local aggregation of the shadow-free proposal with the latent features of the source image.
In this way, it can selectively incorporate the recovered content into the target shadow region while preserving the content of non-target regions. 
By combining these two components, FreeDeshadow enables object-specific shadow removal guided by user prompts, achieving both high-quality restoration and precise object-level control.

To support object-aware controllable shadow removal, we construct PCSRD+, a large-scale dataset for training and evaluation.
Existing shadow removal datasets primarily provide image-level (global) shadow masks and lack object-level prompts and annotations required to remove a specific object's shadow according to a given control prompt.
To resolve this, we annotate each sample in PCSRD+ with six prompt types that identify the target object. 
Three of them are visual prompts based on a dot, a line, and an object mask, which specify the target object with increasing levels of spatial support. 
The other three are textual prompts based on object attributes, spatial relations, and states, respectively, which identify the target object through object-aware reasoning rather than explicit spatial specification.
After careful construction and filtering, the final dataset contains $154,571$ quadruples that include shadow images, user prompts, the shadow masks of the specified objects, and the corresponding shadow-free images.
Notably, the images are provided at a resolution of $1024\times1024$, offering richer texture and structure detail than previous datasets and thus better supporting the training and evaluation of high-quality shadow removal.
We believe the PCSRD+ will facilitate further research on controllable shadow removal and object-aware reasoning.

Our main contributions can be summarized as follows:
\begin{itemize}
\item We propose an object-aware controllable shadow removal paradigm that allows users to specify the target object through either visual or textual prompts and removes only its cast shadow. 
To the best of our knowledge, this is the first work to support both visual and textual prompts within a unified framework for shadow removal.
\item We develop FreeDeshadow, a unified framework that consists of an object-aware generator and a shadow-aware local refiner.
The former produces a high-quality shadow removal proposal conditioned on the user-specified object, while the latter establishes the object-shadow correspondence and confines the generated changes to the target shadow region.
\item We construct PCSRD+, a large-scale controllable shadow removal dataset containing $154,571$ quadruples at $1024\times1024$ resolution and covering six prompt types, including three visual and three textual prompts.
\end{itemize}

Compared with the conference version~\cite{chen2025promptaware}, the additional contributions of this paper are as follows:
\ding{182} We develop a unified and controllable shadow removal framework that extends user prompts from the visual modality alone to both visual and textual modalities.
\ding{183} We add three types of textual prompts to the PCSRD+ dataset, including object attribute reasoning, spatial relation reasoning, and state reasoning.
\ding{184} We expand the dataset from $11,900$ to $154,571$ quadruples and increase the resolution from $256\times256$ to $1024\times1024$, providing more data and finer textures than the previous PCSRD.
\ding{185} We conduct extensive experiments demonstrating that our model achieves state-of-the-art shadow removal performance, while further confirming its superior controllability.

The remainder of this paper is organized as follows. 
Related work on shadow removal, shadow detection, and image editing is reviewed in Section~\ref{sec:related_work}.
Section~\ref{sec:methodology} then details the proposed unified and controllable shadow removal framework, while Section~\ref{sec:shadow_reasoning_dataset} turns to the construction of the PCSRD+ dataset. 
After presenting the experimental results and analysis in Section~\ref{sec:experiments}, we conclude the paper in Section~\ref{sec:conclusion}.

\section{Related Work}
\label{sec:related_work}

\subsection{Shadow Removal}
Shadow removal~\cite{zhang2015shadow,yang2012shadow,guo2012paired,le2019shadowdecomposition,zheng2025shadow,chen2025promptaware} aims to recover the appearance of shadow regions while preserving the original scene structure.
Early methods~\cite{zhang2015shadow,yang2012shadow,guo2012paired} relied on handcrafted illumination, color, or region priors.
Although these priors are interpretable, they often struggle with complex scenes, where shadow appearance is jointly determined by material, geometry, and high-level semantics.
With the development of deep learning, shadow removal has increasingly been formulated as a data-driven image restoration problem. 
DeshadowNet~\cite{qu2017deshadownet} introduced multi-context features for single-image shadow removal, while ST-CGAN~\cite{wang2018stcgan} jointly learned shadow detection and removal with adversarial supervision.
Subsequently, researchers explored shadow decomposition~\cite{le2019shadowdecomposition,zheng2025shadow}, global context modeling~\cite{guo2023shadowformer,liu2024regional}, Transformer architectures~\cite{xiao2024homoformer,dong2025retinex}, and diffusion priors~\cite{xu2025detail,luo2025diff} to improve color consistency and texture recovery.
From a methodological perspective, these deep shadow removal methods can be broadly divided into two categories: shadow-unaware and shadow-aware methods.
Shadow-unaware methods~\cite{wang2018stcgan,yu2025portrait,liu2023shadow} directly predict shadow-free images without requiring explicit shadow masks.
However, without explicit shadow location priors, these methods struggle to distinguish shadows from inherently dark surfaces, resulting in degraded performance in complex scenes.
In contrast, shadow-aware methods~\cite{niu2023boundary,mei2024latent,guo2023shadowformer} introduce shadow-region priors to guide restoration. 
Although such guidance improves the localization of shadow regions, it does not explicitly establish the correspondence between a user-specified object and its cast shadow. 
Consequently, these methods remain limited in providing object-level control, particularly in scenes containing multiple objects and overlapping shadows.

To enable object-level control, our earlier work~\cite{chen2025promptaware} introduced a prompt-aware controllable shadow removal framework.
It allows users to specify target objects through visual prompts, including dots, lines, and object masks, without directly providing target shadow masks.
Building on this, we further extend object-aware controllable shadow removal from visual prompts to both visual and textual prompts. 
Unlike visual prompts that explicitly indicate the target object through spatial cues, textual prompts require the model to interpret object attributes, spatial relations, or states to identify the intended object and establish its correspondence with the cast shadow.
Accordingly, the proposed FreeDeshadow framework integrates multimodal object understanding with object–shadow correspondence modeling to selectively remove the target shadow while preserving other content.

\subsection{Shadow Detection}
Shadow detection~\cite{guo2012paired,wang2018stcgan} aims to localize shadow regions in an image and is closely related to shadow removal.
Existing methods can generally be divided into two categories: generic shadow detection and instance shadow detection. 
Generic shadow detection methods~\cite{guo2012paired,wang2018stcgan,zhu2022single,hu2018direction,zhu2018bidirectional,chen2020multi} classify each pixel as either shadow or non-shadow. 
For example, DSC~\cite{hu2018direction} improved shadow boundary discrimination through a direction-aware spatial context module. 
To better handle shadows on different scales, BDRAR~\cite{zhu2018bidirectional} introduced a bidirectional feature pyramid to propagate and refine features across scales. 
To further improve generalization, MTMT-Net~\cite{chen2020multi} adopted a semi-supervised multi-task learning framework that jointly optimizes several related tasks and leverages unlabeled data through a mean-teacher mechanism.
However, these methods primarily formulate shadow detection as pixel-level binary classification, without distinguishing shadows cast by different objects or establishing object–shadow correspondence~\cite{hu2026unveiling,wang2025metashadow}.
Consequently, they cannot directly identify the cast shadow of a user-specified object, particularly in scenes containing multiple objects or overlapping shadows.

Unlike generic shadow detection, instance shadow detection methods~\cite{wang2020instance,xing2024video,wang2022instance,wang2025metashadow} associate each shadow instance with its corresponding object instance, enabling object-level shadow understanding.
Wang et al.~\cite{wang2020instance} first defined this task and introduced a benchmark dataset for instance shadow detection.
Subsequently, SSIS~\cite{wang2021single} unified object detection, shadow detection, and instance association within a single-stage end-to-end framework to improve inference efficiency. 
Recently, MetaShadow~\cite{wang2025metashadow} introduced an object-centered framework that detects the shadow instance corresponding to a specified object, further extending instance-level shadow understanding.
Despite these advances, these methods either detect all object–shadow pairs without target selection or require labor-intensive precise object masks to specify the target.
Inspired by interactive segmentation~\cite{lai2024lisa,ravi2025sam}, we extend target specification beyond precise masks to more flexible visual and textual prompts.
Our method further integrates multimodal object understanding with object–shadow correspondence modeling to enable selective removal of the target shadow while preserving non-target content.

\begin{figure*}[!t]
\centering
\includegraphics[width=\textwidth]{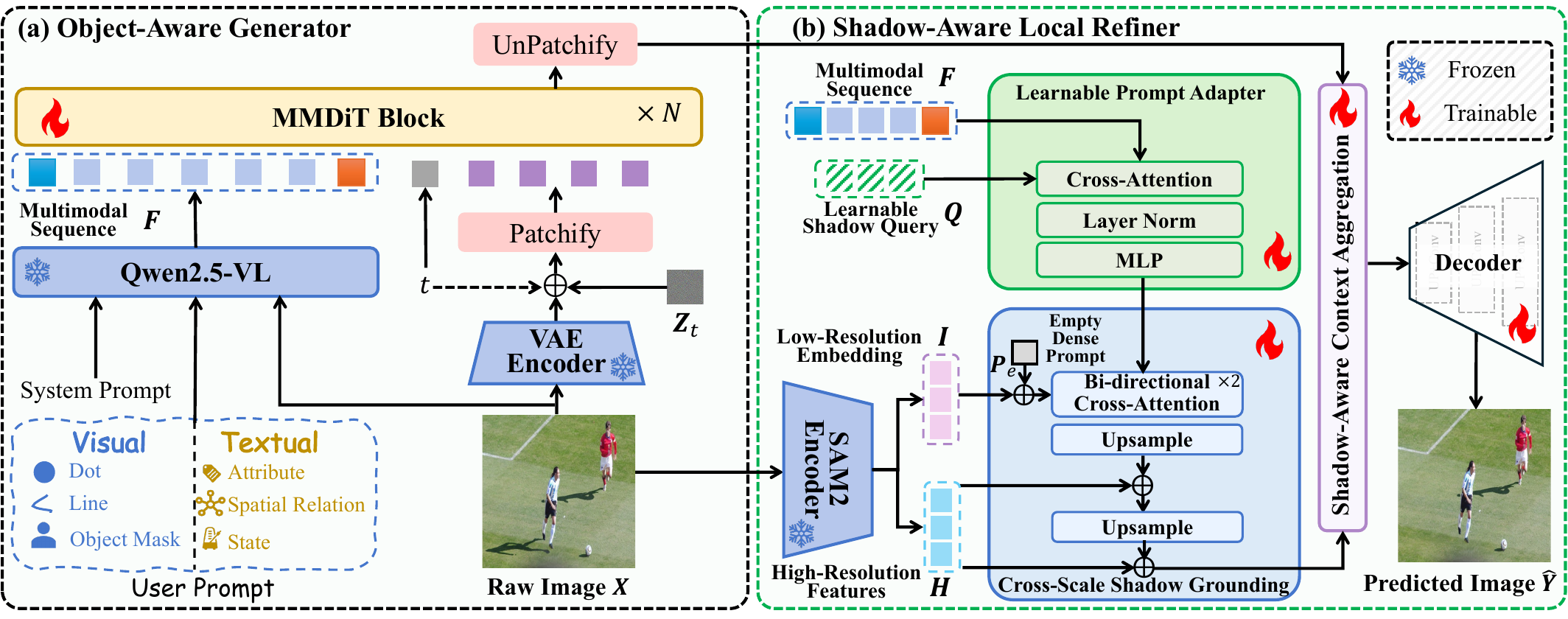}
\caption{Overview of the FreeDeshadow framework. 
(a) The object-aware generator employs Qwen2.5-VL to extract multimodal features from the input image and user prompt, and leverages pretrained generative priors to produce a high-quality shadow-free proposal. 
(b) The shadow-aware local refiner transforms the multimodal features into sparse shadow prompts and employs cross-scale feature interactions to establish the correspondence between the specified object and its cast shadow. 
Guided by this correspondence, the refiner aggregates the shadow-free proposal with the latent features of the source image, thereby constraining the generated changes to the target shadow region while preserving non-target content.
}
\label{fig:framework}
\end{figure*}

\subsection{Image Editing}
Image editing methods~\cite{huang2025diffusion,avrahami2025stable,ye2026imgedit,zhang2026affective,he2026freeedit,li2026brushedit,plitsis2026disentangling} have enabled flexible modification of visual content under textual, spatial, or region-level guidance.
More recently, diffusion and flow-matching models~\cite{liu2025step1xedit,zhang2025icedit,paul2025fluxedit,wu2025qwenimage,mao2025aceplusplus,li2026brushedit} have further advanced image editing by leveraging pretrained generative priors for high-quality visual synthesis.
Instruction-following methods~\cite{brooks2023instructpix2pix,zhao2024ultraedit} enable image modifications through natural language commands, while controllable generation frameworks~\cite{zhang2023controlnet,mao2025aceplusplus} incorporate structural conditions such as masks, edges, and layouts to provide more precise editing control. 
Region-specific editing methods~\cite{zhou2026refineanything,he2026freeedit} further constrain modifications to user-specified regions while preserving non-target content.

Despite these advances, directly applying existing image editing models to object-aware controllable shadow removal remains challenging due to object–shadow asymmetry. 
Specifically, these models struggle to establish the correspondence between the specified object and its spatially separated cast shadow, while confining modifications to the target shadow region.
In this paper, our method leverages pretrained generative priors to recover shadow content and explicitly models object–shadow correspondence to confine restoration to the target shadow region.

\section{Methodology}
\label{sec:methodology}

\subsection{Problem Formulation and Overview}
\label{Overview}

Let $\textbf{\emph{X}}\in\mathbb{R}^{{H}\times{W}\times{3}}$ denote a shadow image. 
Object-aware controllable shadow removal aims to remove the shadow cast by a user-specified object according to a prompt $\mathcal{C}$ and generate an output image $\widehat{\textbf{\emph{Y}}}\in\mathbb{R}^{{H}\times{W}\times{3}}$. 
$\widehat{\textbf{\emph{Y}}}$ should be consistent with the ground-truth image ${\textbf{\emph{Y}}}\in\mathbb{R}^{{H}\times{W}\times{3}}$, \ie, it restores normal illumination within the shadow region of the object specified by $\mathcal{C}$, while preserving the object itself, the shadows of other objects, and the remaining content.
Notably, in our setting, the prompt $\mathcal{C}$ can take various forms, such as a visual prompt (\eg, a dot, a line, or an object mask) or a textual prompt based on attributes, spatial relations, or states.
In this design, the object-aware controllable shadow removal paradigm offers stronger controllability and broader applicability. 

In this paper, we propose a unified framework termed FreeDeshadow to address this task.
As illustrated in Fig.~\ref{fig:framework}, FreeDeshadow consists of two components: an object-aware generator (Section~\ref{sec:oag}) and a shadow-aware local refiner (Section~\ref{Shadow-Aware Local Refiner}).
The object-aware generator employs a frozen Qwen2.5-VL~\cite{bai2025qwen25vl} encoder for multimodal object understanding and leverages pretrained generative priors to produce a high-quality shadow-free proposal conditioned on the user-specified object.
The shadow-aware local refiner transforms the multimodal representation into sparse shadow prompts and employs cross-scale feature interactions to establish the correspondence between the specified object and its cast shadow.
In this way, the model can constrain the generated changes to the target shadow region, while preserving other content.

\subsection{Object-Aware Generator}
\label{sec:oag}
Compared with conventional shadow removal, object-aware controllable shadow removal requires accurately understanding the user prompt to identify the intended object.
However, this is non-trivial, as prompts take diverse forms, including visual prompts such as a dot, a line, or an object mask, and textual prompts based on object attributes, spatial relations, and states. 
Therefore, the model must interpret these heterogeneous prompts to obtain a unified semantic representation of the specified object.
Furthermore, recovering the textures and illumination obscured by shadows requires strong generative priors, which are difficult to learn from scratch due to the scarcity of paired shadow/shadow-free data.
To address these challenges, we develop an object-aware generator based on Qwen-Image-Edit~\cite{wu2025qwenimage}.
It consists of three components.
First, a frozen Qwen2.5-VL~\cite{bai2025qwen25vl} encoder jointly processes the input image and user prompt to extract a unified multimodal representation of the specified object. 
Second, a VAE encoder maps the input image into a latent representation that preserves its spatial structure and appearance.
Third, an MMDiT-based diffusion backbone leverages pretrained generative priors to produce a high-quality shadow-free proposal conditioned on the multimodal representation and image latent.

Specifically, we first employ a frozen Qwen2.5-VL~\cite{bai2025qwen25vl} encoder to map the shadow image $\textbf{\emph{X}}$ and the user prompt $\mathcal{C}$ into a full multimodal hidden sequence ${\textbf{\emph{F}}}$.
Then, the shadow image $\textbf{\emph{X}}$ is encoded into the latent space via a VAE encoder.
\begin{equation}
\begin{gathered}
{\textbf{\emph{F}}}
=
\mathcal{E}_{\texttt{Qwen}}(\textbf{\emph{X}},\mathcal{C}),
~~
{\textbf{\emph{E}}}
=
\mathcal{V}_{\texttt{VAE}}(\textbf{\emph{X}}),
\end{gathered}
\label{Generator}
\end{equation}
where ${\textbf{\emph{F}}}\in\mathbb{R}^{L\times d}$ denotes the full multimodal hidden sequence with sequence length $L$ and feature dimension $d$.
${\textbf{\emph{E}}}$ is the latent representation of the shadow image $\textbf{\emph{X}}$.
$\mathcal{E}_{\texttt{Qwen}}(\cdot,\cdot)$ and $\mathcal{V}_{\texttt{VAE}}(\cdot)$ denote the frozen Qwen2.5-VL~\cite{bai2025qwen25vl} encoder and the VAE encoder, respectively.
In this way, ${\textbf{\emph{F}}}$ captures the high-level semantics of the target object and the user's removal intent to condition the MMDiT backbone through joint attention~\cite{yang2025cogvideox}.
It also retains the full multimodal context for the shadow-aware local refiner.
Meanwhile, the latent representation ${\textbf{\emph{E}}}$ preserves the spatial structure and appearance of the shadow image $\textbf{\emph{X}}$, providing image-aligned visual conditioning.

After obtaining the latent representation ${\textbf{\emph{E}}}$, we concatenate it with the noisy latent ${\textbf{\emph{Z}}}_t$ and patchify the resulting representation into image tokens.
These tokens are fed into the MMDiT denoiser from Qwen-Image-Edit-2511~\cite{wu2025qwenimage} to iteratively denoise ${\textbf{\emph{Z}}}_t$.
To adapt the model for shadow removal, we equip the MMDiT attention projections with trainable low-rank adaptation (LoRA) adapters~\cite{hu2022lora}.
During inference, given a shadow image $\textbf{\emph{X}}$ and a user prompt $\mathcal{C}$, we initialize ${\textbf{\emph{Z}}}_T$ with Gaussian noise and iteratively denoise it according to the diffusion schedule to obtain ${\textbf{\emph{Z}}}_0$.
We then unpatchify it into the shadow-free proposal $\textbf{\emph{S}}$.
In this way, the object-aware generator combines multimodal semantic conditioning with image-aligned latent guidance to provide a high-quality proposal for subsequent local refinement.

\subsection{Shadow-Aware Local Refiner}
\label{Shadow-Aware Local Refiner}
The object-aware generator produces a high-quality shadow-free proposal through a global diffusion process.
However, this process may introduce unintended changes outside the target shadow region. 
Moreover, due to object–shadow asymmetry, the specified object and its cast shadow often occupy different spatial regions and exhibit distinct appearances, making it challenging to identify the target shadow from the object representation alone.
To address this, we design a shadow-aware local refiner. 
It establishes the correspondence between the object specified by the user prompt and its cast shadow, and leverages this correspondence to confine the refinement of the generator's global proposal to the target shadow region and its boundary neighborhood.
In this way, the refiner preserves non-target region content while focusing the refinement on the target shadow region.
The refiner consists of a learnable prompt adapter, a cross-scale shadow grounding module, and a shadow-aware context aggregation module.

\noindent\textbf{Learnable Prompt Adapter.}
The multimodal hidden sequence ${\textbf{\emph{F}}}$ captures the semantics of the user-specified object and its contextual information.
However, due to object–shadow asymmetry, the object representation cannot directly indicate the spatial location of its cast shadow.
Therefore, we introduce a learnable prompt adapter to map ${\textbf{\emph{F}}}$ into sparse shadow prompts for subsequent cross-scale feature interactions.

Specifically, we first introduce the learnable shadow query tokens ${\textbf{\emph{Q}}}\in\sR^{32\times d}$, and feed them with the full multimodal hidden sequence ${\textbf{\emph{F}}}$ 
into a learnable prompt adapter.
Taking ${\textbf{\emph{Q}}}$ as queries and ${\textbf{\emph{F}}}$ as keys and values, the adapter aggregates object-relevant information through cross-attention and projects the aggregated features into sparse shadow prompts.
In this way, the object representation is transformed into a sparse shadow prompt ${\textbf{\emph{P}}}$ for subsequent cross-scale shadow grounding.
This process is formulated as
\begin{equation}
{\textbf{\emph{P}}}
=
\texttt{Adapter}\!\left(
{\textbf{\emph{Q}}},
{\textbf{\emph{F}}} 
\right),
\label{eq:prompt_adapter}
\end{equation}
where $\texttt{Adapter}(\cdot,\cdot)$ denotes the learnable prompt adapter. 

\noindent\textbf{Cross-Scale Shadow Grounding.}
Shadows often exhibit soft and diffuse boundaries, making it difficult for single-scale features to support both accurate semantic localization and precise boundary delineation.
To address this issue, we employ the frozen SAM2~\cite{ravi2025sam} image encoder to map the shadow image ${\textbf{\emph{X}}}$ into a low-resolution image embedding ${\textbf{\emph{I}}}$ and the hierarchical high-resolution features ${\textbf{\emph{H}}}$.
The former captures global semantic context, while the latter preserves fine-grained spatial details.
This encoding process is formulated as
\begin{equation}
{\textbf{\emph{I}}},
{\textbf{\emph{H}}}
=
\mathcal{E}_{\texttt{SAM2}}({\textbf{\emph{X}}}),
\label{eq:sam_encoder}
\end{equation}
where $\mathcal{E}_{\texttt{SAM2}}(\cdot)$ denotes the image encoder of SAM2~\cite{ravi2025sam}.

Subsequently, the sparse shadow prompt ${\textbf{\emph{P}}}$ is concatenated with the learnable SAM2 output tokens ${\textbf{\emph{U}}}$ to form the prompt sequence ${\textbf{\emph{T}}}_{0}=[{\textbf{\emph{U}}},{\textbf{\emph{P}}}]$.
Meanwhile, the learnable empty dense prompt ${\textbf{\emph{P}}}_{e}$ is added to the image embedding ${\textbf{\emph{I}}}$, yielding ${\textbf{\emph{V}}}_{0}={\textbf{\emph{P}}}_{e}+{\textbf{\emph{I}}}$. 
Based on these representations, we employ two layers of bidirectional cross-attention to enable interaction between prompt and visual features, progressively associating the object semantics encoded in the prompt with the corresponding cast shadow while suppressing irrelevant regions.
\begin{equation}
{\textbf{\emph{T}}}_{l},{\textbf{\emph{V}}}_{l}=\mathcal{BCA}\!\left({\textbf{\emph{T}}}_{l-1},{\textbf{\emph{V}}}_{l-1}\right),\qquad l=1,2,
\label{eq:bidirectional_cross_attention}
\end{equation}
where $\mathcal{BCA}(\cdot,\cdot)$ denotes the bidirectional cross-attention operation.
Finally, the updated visual feature ${\textbf{\emph{V}}}_{2}$ is upsampled and fused with the hierarchical high-resolution features ${\textbf{\emph{H}}}$ to capture fine-grained shadow boundaries and local structures, producing the object-aware shadow feature ${\textbf{\emph{G}}}$.

\noindent\textbf{Shadow-Aware Context Aggregation.}
After obtaining the object-aware shadow feature ${\textbf{\emph{G}}}$, we further constrain the global candidate generated by the generator to the target shadow region associated with the user-specified object.
%we further constrain the global candidate generated by the generator to the region of user-specified object.
Specifically, we concatenate the latent representation of the shadow image $\textbf{\emph{X}}$, the representation $\textbf{\emph{S}}$ of the object-aware generator, and the feature ${\textbf{\emph{G}}}$ along the channel dimension.
The resulting features are then fed into the shadow-aware context aggregation module.
Since ${\textbf{\emph{G}}}$ captures the correspondence between the user-specified object and its cast shadow, it provides spatial guidance for feature aggregation.
Under this guidance, the module adaptively integrates representation $\textbf{\emph{S}}$ of the object-aware generator with the latent representation of the shadow image $\textbf{\emph{X}}$, and activates the candidate features only in regions related to the target shadow.
In this way, the generator output is restricted to the shadow region of the user-specified object, enabling spatially selective local editing.

\subsection{Training Objectives}
As described in Section~\ref{Overview}, the FreeDeshadow framework consists of two components: an object-aware generator and a shadow-aware local refiner.
The former performs global coarse-grained shadow removal, while the latter focuses on fine-grained local refinement of the shadow regions of the user-specified object.
In this setting, the optimization objective of FreeDeshadow consists of the following two terms.

For the object-aware generator, we fine-tune Qwen-Image-Edit~\cite{wu2025qwenimage} using a flow-matching objective to learn shadow removal conditioned on both the input image and the user prompt.
Given a noisy image latent ${\textbf{\emph{Z}}}_t$ at timestep $t$, the object-aware generator learns to predict ${\textbf{\emph{V}}}_{\theta}({\textbf{\emph{Z}}}_t,t,{\textbf{\emph{F}}},{\textbf{\emph{E}}})$ under the condition of the shadow image latent ${\textbf{\emph{E}}}$ and multimodal hidden sequence ${\textbf{\emph{F}}}$.
The training objective of the object-aware generator can be denoted as:
\begin{equation}
\mathbb{E}_{{\textbf{\emph{Z}}}_t,t,{\textbf{\emph{F}}},{\textbf{\emph{E}}}}\Big[\omega_t\parallel{{\textbf{\emph{V}}}_{\theta}({\textbf{\emph{Z}}}_t,t,{\textbf{\emph{F}}},{\textbf{\emph{E}}})-{\textbf{\emph{V}}}_{t}}\parallel^2_2\Big],
\end{equation}
where $\textbf{\emph{V}}_{\theta}(\cdot)$ denotes the MMDiT flow predictor.
${\textbf{\emph{V}}}_{t}$ and $\omega_t$ are the target flow and the timestep weight, respectively.

For the shadow-aware local refiner, we adopt the $\ell_1$-norm to supervise the pixel-level reconstruction accuracy of the results, and we decompose it into the target shadow regions and non-target regions.
The training objective of the shadow-aware local refiner can be denoted as:
\begin{equation}
\mathbb{E}_{\textbf{\emph{Y}},\textbf{\emph{M}}}
\left[
\frac{
\left\|
\textbf{\emph{M}}\odot
(\widehat{\textbf{\emph{Y}}}-\textbf{\emph{Y}})
\right\|_1
}{
\left\|\textbf{\emph{M}}\right\|_1+\epsilon
}
+
\lambda
\frac{
\left\|
(1-\textbf{\emph{M}})\odot
(\widehat{\textbf{\emph{Y}}}-\textbf{\emph{Y}})
\right\|_1
}{
\left\|1-\textbf{\emph{M}}\right\|_1+\epsilon
}
\right],
\label{eq:local_refinement_loss}
\end{equation}
where $\odot$ denotes the Hadamard product.
$\textbf{\emph{M}}$ and $\lambda$ are the mask of shadow regions and the trade-off parameter, respectively.

\section{Shadow Reasoning Dataset}
\label{sec:shadow_reasoning_dataset}

\begin{table*}[!t]
\centering
\caption{Comparison with existing datasets. 
S-mask: shadow mask; 
Obj-S: object-shadow association; Obj-mask: object mask; 
Task: SR, SG, and CSR denote shadow removal, shadow generation, and controllable shadow removal, respectively.}
\vspace{-0.2cm}
\label{tab:dataset_comparison}
\setlength{\tabcolsep}{7pt}
\renewcommand{\arraystretch}{1.0}
\resizebox{1.0\linewidth}{!}{
\begin{tabular}{|l|c|c|c|ccc|ccc|ccc|}
\hline\thickhline
\rowcolor{gray!25}
&  &  &  &
\multicolumn{3}{c|}{Annotation} & \multicolumn{3}{c|}{Visual Prompt} & \multicolumn{3}{c|}{Textual Prompt} \\
\rowcolor{gray!25}
\multirow{-2}{*}{Dataset} & \multirow{-2}{*}{Samples} & \multirow{-2}{*}{Resolution} & \multirow{-2}{*}{Task} & Pair & S-mask & Obj-S & Dot & Line & Obj-mask & V-attr. & Spatial & State \\
\hline
ISTD~\cite{wang2018stcgan} & 1,870 & $480{\times}640$ & SR & \cmark & \cmark & \xmark & \xmark & \xmark & \xmark & \xmark & \xmark & \xmark \\
ISTD+~\cite{le2019shadowdecomposition} & 1,870 & $480{\times}640$ & SR & \cmark & \cmark & \xmark & \xmark & \xmark & \xmark & \xmark & \xmark & \xmark \\
SRD~\cite{qu2017deshadownet} & 3,088 & $640{\times}480$ & SR & \cmark & \xmark & \xmark & \xmark & \xmark & \xmark & \xmark & \xmark & \xmark \\
USR~\cite{hu2019maskshadowgan} & 4,215 & $400{\times}400$ & SR & \xmark &  \xmark & \xmark & \xmark & \xmark & \xmark & \xmark & \xmark & \xmark \\
DESOBA~\cite{hong2021shadow} & 3,623 & $256{\times}256$ & SG & \cmark & \cmark & \cmark & \xmark & \xmark & \cmark & \xmark & \xmark & \xmark \\ 
DESOBAv2~\cite{liu2024shadowgeneration} & 28,573 & $1024{\times}1024$ & SG & \cmark & \cmark & \cmark & \xmark & \xmark & \cmark & \xmark & \xmark & \xmark \\
PCSRD~\cite{chen2025promptaware} & 11,900 & $256{\times}256$ & CSR & \cmark & \cmark & \cmark & \cmark & \cmark & \cmark & \xmark & \xmark & \xmark \\  \hline
\rowcolor{gray!10} \textbf{PCSRD+ (Ours)} & 154,571 & $1024{\times}1024$ & CSR & \cmark & \cmark & \cmark & \cmark & \cmark & \cmark & \cmark & \cmark & \cmark \\
\hline
\end{tabular}}
\end{table*}

\begin{figure}[t!]
\centering
\includegraphics[width=1\linewidth]{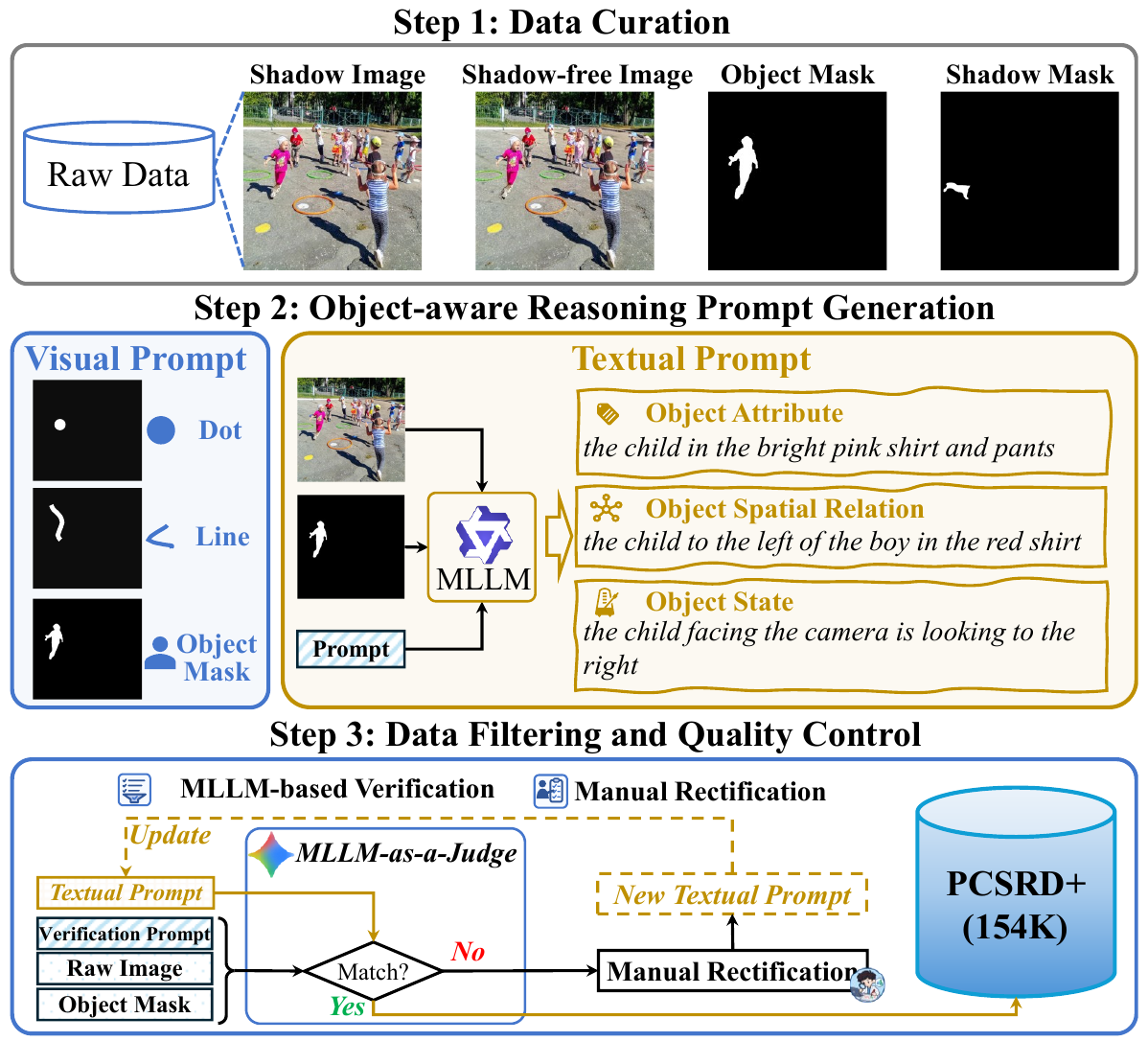}
\vspace{-0.6cm}
\caption{Pipeline for the construction of PCSRD+.}
\label{fig:data_construction}
\vspace{-0.1cm}
\end{figure}

As introduced in Section~\ref{sec:methodology}, the proposed method requires visual or textual prompts that identify the target object. 
As shown in Table~\ref{tab:dataset_comparison}, existing shadow removal datasets~\cite{wang2018stcgan,le2019shadowdecomposition,qu2017deshadownet,hu2019maskshadowgan} do not jointly provide diverse visual prompts and reasoning-aware textual prompts. 
To bridge this gap, we construct an object-aware controllable shadow removal dataset, termed PCSRD+, 
which provides six types of visual and textual prompts. 
PCSRD+ is built upon the DESOBAv2~\cite{liu2024shadowgeneration} dataset, due to its large scale and high-quality annotations. 
The construction pipeline of PCSRD+ is illustrated in Fig.~\ref{fig:data_construction}.

\subsection{Data Curation}
A shadow mask specifies the spatial support of a shadow, but does not by itself provide the object-level information required for controllable shadow removal. In particular, our goal is not merely to learn a mapping from a shadow image to its shadow-free counterpart, but to determine \emph{which object's shadow} should be removed according to a given control cue. 
Motivated by this requirement, we construct PCSRD+ from the $28,573$ object-shadow pairs released with DESOBAv2~\cite{liu2024shadowgeneration}.
Rather than treating each cue-conditioned sample as an independent example, we organize the dataset around a set of base pairs indexed by $\mathcal{I}$.
Each base tuple is defined as
\begin{equation}
    b_i = (\mX_i, \mY_i, \mO_i, \mM_i), \qquad i \in \mathcal{I},
\end{equation}
where $\mX_i$, $\mY_i$, and $\mM_i$ are the sample-specific instances of $\mX$, $\mY$, and $\mM$ introduced in Section~\ref{sec:methodology}, and $\mO_i$ denotes the auxiliary mask of the object. 
We follow the official DESOBAv2 split~\cite{liu2024shadowgeneration}, with $27,823$ base tuples for training and $750$ for testing. 
All cue variants generated from the same DESOBAv2 pair inherit its split to avoid data leakage.

\subsection{Reasoning-aware Prompt Generation}
For each base tuple, we construct prompts from the following six types:
\begin{equation}
    \mathcal{C}
    =
    \{
        \mathcal{C}_p,
        \mathcal{C}_l,
        \mathcal{C}_m,
        \mathcal{C}_a,
        \mathcal{C}_r,
        \mathcal{C}_s
    \}.
\end{equation}
Here, $\mathcal{C}_p$, $\mathcal{C}_l$, and $\mathcal{C}_m$ denote dot, line, and object mask prompts, respectively. They follow the visual prompt setting in our previous work~\cite{chen2025promptaware} and specify the target object through spatial information with different levels of granularity. 
In contrast, $\mathcal{C}_a$, $\mathcal{C}_r$, and $\mathcal{C}_s$ denote attribute-based, spatial-relation-based, and state-based textual prompts, respectively.
These textual prompts are instantiated as reasoning-oriented referring expressions. Rather than explicitly indicating the target location, they require the model to identify the object from its appearance, its relationship with other scene elements, or its visible state.

For an applicable prompt type $\tau\in\mathcal{C}_i$, the corresponding prompt-conditioned sample is defined as
\begin{equation}
    r_i^{\tau}
    =
    \left(
        \mX_i,
        c_i^{\tau},
        \mY_i,
        \mM_i
    \right),
\end{equation}
where $c_i^{\tau}$ is the prompt and $\mathcal{C}_i\subseteq\mathcal{C}$ contains the prompt types applicable to base tuple $i$. 
For visual prompts, $c_i^{\tau}$ is provided as an additional prompt image. For textual prompts, it is incorporated into the editing instruction as a referring expression. Importantly, none of the six prompt types directly specifies the target-shadow mask $\mM_i$.

\subsubsection{Visual Prompts}
To support controllable shadow removal without using the object-shadow mask as a control input, we use visual prompts to specify the object whose cast shadow should be removed. 
These prompts identify the relevant object in the input image rather than directly marking its shadow. We use three forms of visual control: a dot, a line, and an object mask. 
The dot and line serve as sparse indicators of the object, while the object mask provides its complete annotated region. Each cue is constructed from the same object annotation, ensuring that the target object remains consistent across cue types. 
Given an object mask, we derive three visual prompts with different levels of spatial support for specifying the target object. 
\textbf{(i)} The \textit{dot cue} is centered at the maximally interior location of the mask, as determined by the Euclidean distance transform.  
\textbf{(ii)} The \textit{line cue} is constructed from the central portion of an approximate longest geodesic path along the skeleton of the mask’s largest connected component. When a valid skeleton path cannot be obtained, an in-mask principal-axis segment is used instead.
\textbf{(iii)} The \textit{object mask cue} retains the complete object region. 
All three prompts are derived exclusively from the object annotation and therefore specify the target object without revealing its shadow mask.

\subsubsection{Textual Prompts}
Textual prompts are introduced to enable controllable shadow removal through object-aware reasoning rather than explicit spatial specification. Unlike dot, line, and object-mask prompts, a textual cue does not directly indicate the location or spatial extent of the target object. 
The model must first ground the referring expression to the correct object in the image and then associate that object with its cast shadow. 
Textual control therefore requires both language-based object grounding and object-to-shadow reasoning. 
Specifically, we define three reasoning categories according to the visual evidence required to identify the target object. 
\textbf{(i)} \textit{Object Attribute Reasoning} requires matching the referring expression to intrinsic appearance, such as clothing, color, or other visible characteristics. 
\textbf{(ii)} \textit{Object Spatial Relation Reasoning} requires resolving the object’s position relative to visible objects or regions in the scene. 
\textbf{(iii)} \textit{Object State Reasoning} requires recognizing pose, orientation, physical state, or direct interaction with another object. 
A reasoning category is considered applicable only when its corresponding evidence uniquely distinguishes the target object from other objects in the image.

Following previous work on language-guided visual reasoning~\cite{yue2025mmmu,wang2025visualprm}, we adopt a prompt-based VLM annotation pipeline to construct reasoning-oriented textual prompts. 
For each sample, Qwen3.6-27B~\cite{qwen3.6-27b} is provided with the input image, the object mask used solely to specify the annotation target, and the object description released with DESOBAv2 when available. 
Under category-specific instructions, the VLM determines whether the required visual evidence can uniquely identify the target object. It then returns an eligibility decision, a concise referring expression, a confidence score, and a brief rationale in a structured response.
The three tasks require distinct evidence for target identification: intrinsic appearance for $\mathcal{C}_a$, relations to visible scene anchors for $\mathcal{C}_r$, and visible pose, orientation, physical state, or direct contact for $\mathcal{C}_s$.

\begin{figure}[!t]
\centering
\includegraphics[width=1\linewidth]{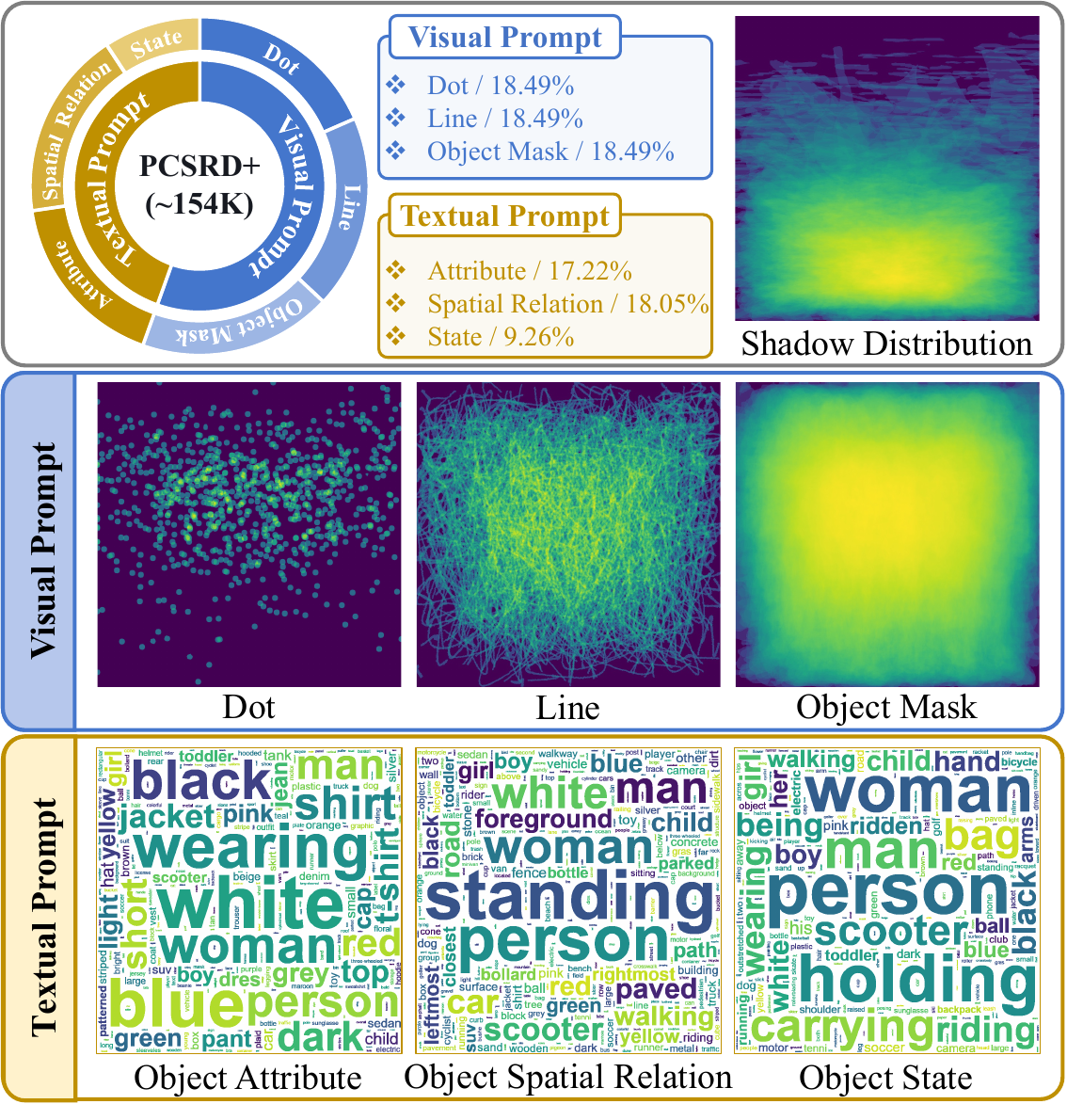}
\vspace{-1.8em}
\caption{Statistics for the PCSRD+ dataset.}
\label{fig:dataset_statistics}
\end{figure}

\subsection{Data Filtering and Quality Control}

VLM-generated referring expressions may be ambiguous, insufficiently discriminative, or contaminated by annotation-specific artifacts. To improve annotation reliability, we employ a two-stage quality-control procedure consisting of MLLM-based filtering and human verification.

\subsubsection{MLLM-based Verification} 
For each sample, we provide an MLLM-as-a-Judge~\cite{chen2024mllm} with the generated textual prompt, raw image, object mask, and a predefined verification prompt. 
The MLLM (\textit{i.e.,} Gemini-2.5-Flash~\cite{comanici2025gemini}) determines whether the textual prompt accurately and unambiguously matches the target object indicated by the mask. 
Samples judged to be correctly matched are directly retained in PCSRD+, while mismatched samples are forwarded for manual rectification.

\subsubsection{Human Rectification}
To complement the MLLM-based verification, we introduce human rectification to revise the mismatched samples. 
For each sample rejected by the MLLM verification module, a human annotator inspects the raw image, object mask, and original textual prompt, and then revises the prompt to produce a more accurate and discriminative referring expression. 
The corrected textual prompt is subsequently fed back into the MLLM-based verification step for another round of checking. 
This verification-rectification process is repeated until the textual prompt correctly matches the target object. After the filtering and quality-control procedure, we obtain PCSRD+, containing approximately $154$K validated samples.

\subsection{Data Statistics}
The PCSRD+ dataset is constructed from $28,573$ base pairs, of which $27,823$ are assigned to the training set and $750$ to the test set. 
For the visual cue part, each base pair is paired with three types of prompts, dot, line, and object mask, bringing the total number of samples to $85,719$. 
For the textual cue part, we retain only prompts that uniquely identify a single target object. 
Applying this criterion results in $26,625$ attribute-reasoning samples, $27,909$ spatial-relation samples, and $14,318$ state-configuration samples, for a total of $68,852$ textual-cue samples. 
Overall, the benchmark comprises $154,571$ samples, including $150,516$ training samples and $4,055$ test samples. 

As shown in Fig.~\ref{fig:dataset_statistics}, we summarize the statistical characteristics of PCSRD+ from four perspectives: sample counts, object-category distribution, the spatial distribution of visual prompts, and word clouds of textual prompts. 
For the visual prompts, we can see that shadows occur predominantly in the central and lower regions of the images. Dot prompts are similarly concentrated near the image center, whereas line prompts span a broader spatial range because they are derived from central in-mask skeletal paths or principal-axis segments of the target objects. Object-mask prompts exhibit a relatively smooth central distribution, consistent with the overall object distribution. 
For the textual prompts, attribute-reasoning prompts primarily describe appearance-related properties, such as color and clothing; spatial-relation prompts emphasize relative positions, scene regions, and surrounding objects; and state-configuration prompts focus on actions, poses, object interactions, and physical states. 
Overall, these distributions demonstrate that PCSRD+ provides diverse and complementary prompts for reasoning about target shadows under different forms of spatial and semantic guidance.

\section{Experiments}
\label{sec:experiments}

\begin{table*}[!t]
\centering
\caption{Quantitative comparison on the PCSRD+ dataset.}
\label{tab:main_model_metrics}
\tabcolsep 4pt
\resizebox{1.0\textwidth}{!}{
\begin{tabular}{|r|cccc|cccc|cccc|}
\hline\thickhline
\rowcolor{gray!25}
& \multicolumn{4}{c|}{Shadow Regions} & \multicolumn{4}{c|}{Non-Shadow Regions} & \multicolumn{4}{c|}{All Regions} \\
\rowcolor{gray!25}
\multirow{-2}{*}{Method} & PSNR $\uparrow$ & SSIM $\uparrow$ & RMSE $\downarrow$ & LPIPS $\downarrow$ & PSNR $\uparrow$ & SSIM $\uparrow$ & RMSE $\downarrow$ & LPIPS $\downarrow$ & PSNR $\uparrow$ & SSIM $\uparrow$ & RMSE $\downarrow$ & LPIPS$\downarrow$ \\
\hline
\rowcolor{gray!10}\multicolumn{13}{|l|}{\textbf{Visual Prompts (PCSRD+, $1024\times1024$)}} \\
BMNet~\cite{zhu2022bijective} & 9.885 & 0.3043 & 87.0258 & 0.0281 & 24.350 & 0.8205 & 17.2356 & 0.2826 & 21.951 & 0.8086 & 22.3503 & 0.2980 \\
ControlNet-Inpainting~\cite{zhang2023controlnet} & 10.231 & 0.2802 & 83.4050 & 0.0264 & 25.146 & 0.7059 & 16.5155 & 0.1424 & 22.639 & 0.6969 & 21.3499 & 0.1593 \\
Inpaint4Shadow~\cite{li2023leveraging} & 11.005 & 0.3376 & 77.3426 & 0.0267 & 27.807 & 0.8224 & 11.7897 & 0.2795 & 24.485 & 0.8114 & 17.2013 & 0.2967 \\
ShadowFormer~\cite{guo2023shadowformer} & 10.826 & 0.3340 & 78.7570 & 0.0269 & 28.836 & 0.8296 & 10.6695 & 0.2717 & 24.907 & 0.8183 & 16.5844 & 0.2893 \\
ShadowDiffusion~\cite{guo2023shadowdiffusion} & 10.880 & 0.3369 & 78.0889 & 0.0269 & 28.526 & 0.8289 & 10.8177 & 0.2680 & 24.829 & 0.8177 & 16.5310 & 0.2854 \\
FLUX.1 Fill [dev]~\cite{blackforestlabs2024fluxfilldev} & 10.299 & 0.2774 & 82.5398 & 0.0275 & 31.370 & 0.9531 & 8.7635 & 0.0409 & 25.970 & 0.9382 & 15.5067 & 0.0617 \\
PowerPaint~\cite{zhuang2024powerpaint} & 13.196 & 0.4800 & 59.3231 & 0.0328 & 10.287 & 0.4132 & 79.0009 & 0.8369 & 10.336 & 0.4161 & 78.5448 & 0.8511 \\
HomoFormer~\cite{xiao2024homoformer} & 10.810 & 0.3337 & 78.9105 & 0.0270 & 28.619 & 0.8246 & 10.6606 & 0.2752 & 24.838 & 0.8134 & 16.5138 & 0.2926 \\
FireRed-Image-Edit-1.1~\cite{fireredteam2025imageedit11} & 14.738 & 0.4185 & 56.3747 & 0.0171 & 29.219 & 0.9317 & 10.4019 & 0.0773 & 26.448 & 0.9211 & 13.6644 & 0.0872 \\
FireRed-Image-Edit-1.0~\cite{fireredteam2025imageedit10} & 15.069 & 0.4141 & 55.4391 & 0.0154 & 28.271 & 0.8965 & 11.0319 & 0.0852 & 26.281 & 0.8869 & 13.6742 & 0.0948 \\
FLUX.2 [dev]~\cite{blackforestlabs2025flux2dev} & 13.502 & 0.3792 & 65.3681 & 0.0213 & 28.835 & 0.8680 & 10.1379 & 0.1109 & 25.729 & 0.8575 & 14.5737 & 0.1254 \\
ACE++~\cite{mao2025aceplusplus} & 10.487 & 0.2828 & 80.9918 & 0.0274 & 32.877 & 0.9573 & 8.8150 & 0.0387 & 26.372 & 0.9425 & 15.7271 & 0.0598 \\
HD-Painter~\cite{manukyan2025hdpainter} & 10.967 & 0.3243 & 77.3510 & 0.0254 & 24.860 & 0.7189 & 17.1790 & 0.1599 & 22.666 & 0.7106 & 21.4612 & 0.1757 \\
BrushEdit~\cite{li2026brushedit} & 11.251 & 0.3178 & 75.5514 & 0.0275 & 25.859 & 0.7697 & 20.7622 & 0.1759 & 22.152 & 0.7596 & 25.8137 & 0.1937 \\
Qwen-Image-Edit~\cite{wu2025qwenimage} & 10.008 & 0.2603 & 87.8359 & 0.0314 & 23.640 & 0.7161 & 30.8969 & 0.2615 & 20.841 & 0.7060 & 35.6154 & 0.2785 \\
Qwen-Image-Edit-2509~\cite{wu2025qwenimage} & 11.141 & 0.3770 & 78.5781 & 0.0302 & 23.091 & 0.7814 & 38.3210 & 0.2655 & 19.950 & 0.7717 & 42.4577 & 0.2809 \\
Qwen-Image-Edit-2511~\cite{wu2025qwenimage} & 8.661 & 0.2117 & 100.7679 & 0.0332 & 16.126 & 0.5779 & 46.5890 & 0.4113 & 15.277 & 0.5693 & 49.7686 & 0.4228 \\
Qwen-Image-Edit-2511~\cite{wu2025qwenimage} + LoRA & \toptwo{21.857} & \toptwo{0.5672} & \topone{23.4350} & \toptwo{0.0136} & \toptwo{39.553} & \toptwo{0.9739} & \toptwo{2.9837} & \toptwo{0.0234} & \toptwo{34.459} & \toptwo{0.9567} & \toptwo{5.4609} & \toptwo{0.0347} \\ \hline
\rowcolor{gray!10}\textbf{FreeDeshadow (Ours)} & \topone{22.018} & \topone{0.5756} & \toptwo{23.4455} & \topone{0.0090} & \topone{51.078} & \topone{0.9977} & \topone{0.9222} & \topone{0.0037} & \topone{40.323} & \topone{0.9876} & \topone{3.1272} & \topone{0.0112} \\
\hline

\rowcolor{gray!10}\multicolumn{13}{|l|}{\textbf{Textual Prompts (PCSRD+, $1024\times1024$)}} \\
SD1.5 Img2Img~\cite{stabilityai2022sd15} & 11.524 & 0.3635 & 75.0929 & 0.0311 & 17.942 & 0.5373 & 33.0921 & 0.4936 & 17.302 & 0.5326 & 35.6137 & 0.5052 \\
% MagicBrush~\cite{zhang2023magicbrush} & 26.561 & 0.6784 & 16.1595 & 0.0052 & 22.030 & 0.6869 & 22.0444 & 0.1667 & 22.088 & 0.6868 & 21.8817 & 0.1696 \\
InstructPix2Pix~\cite{brooks2023instructpix2pix} & 13.075 & 0.3876 & 63.1173 & 0.0353 & 14.556 & 0.5349 & 50.3386 & 0.6380 & 14.398 & 0.5317 & 51.0177 & 0.6505 \\
SDXL-InstructPix2Pix~\cite{diffusers2023sdxlinstructpix2pix} & 10.225 & 0.3262 & 85.2992 & 0.0369 & 11.165 & 0.4664 & 73.8361 & 0.7096 & 11.066 & 0.4630 & 74.4539 & 0.7211 \\
OmniGen~\cite{xiao2024omnigen} & 17.117 & 0.4911 & 45.2235 & 0.0191 & 25.247 & 0.8304 & 16.3989 & 0.1503 & 24.236 & 0.8221 & 18.0335 & 0.1628 \\
UltraEdit~\cite{zhao2024ultraedit} & 10.893 & 0.2792 & 77.9560 & 0.0265 & 26.282 & 0.8029 & 12.9660 & 0.2129 & 23.784 & 0.7913 & 17.9002 & 0.2281 \\
FireRed-Image-Edit-1.1~\cite{fireredteam2025imageedit11} & 16.031 & 0.4560 & 48.3546 & 0.0163 & 25.751 & 0.8968 & 15.5940 & 0.0962 & 24.408 & 0.8867 & 17.3960 & 0.1044 \\
FireRed-Image-Edit-1.0~\cite{fireredteam2025imageedit10} & 13.984 & 0.4069 & 59.1347 & 0.0190 & 27.797 & 0.9181 & 12.2931 & 0.0759 & 25.208 & 0.9061 & 15.6655 & 0.0869 \\
FLUX.2 [dev]~\cite{blackforestlabs2025flux2dev} & 15.250 & 0.4421 & 55.8048 & 0.0211 & 26.107 & 0.8670 & 14.8123 & 0.1320 & 24.021 & 0.8565 & 18.1035 & 0.1448 \\
Step1X-Edit~\cite{liu2025step1xedit} & 12.780 & 0.3952 & 67.3185 & 0.0250 & 26.245 & 0.8419 & 14.1299 & 0.1268 & 23.593 & 0.8310 & 18.5398 & 0.1399 \\
ICEdit~\cite{zhang2025icedit} & 10.508 & 0.3212 & 80.9713 & 0.0295 & 25.075 & 0.8204 & 17.4484 & 0.1452 & 22.136 & 0.8079 & 22.9463 & 0.1613 \\
ACE++~\cite{mao2025aceplusplus} & 9.878 & 0.3146 & 87.7319 & 0.0426 & 9.075 & 0.3317 & 91.4677 & 0.9794 & 9.069 & 0.3311 & 91.5304 & 0.9982 \\
FluxEdit~\cite{paul2025fluxedit} & 9.791 & 0.2981 & 86.7199 & 0.0410 & 20.934 & 0.8730 & 26.2223 & 0.1318 & 18.733 & 0.8533 & 31.5069 & 0.1589 \\
FLUX.1-Kontext-dev~\cite{blackforestlabs2025fluxkontextdev} & 18.688 & 0.5139 & 38.8582 & 0.0159 & 29.411 & 0.9548 & 11.4293 & 0.0482 & 27.338 & 0.9442 & 13.4358 & 0.0601 \\
BrushEdit~\cite{li2026brushedit} & 13.234 & 0.3765 & 59.7642 & 0.0297 & 15.623 & 0.6014 & 43.7001 & 0.3506 & 15.455 & 0.5961 & 44.4303 & 0.3659 \\
Qwen-Image-Edit~\cite{wu2025qwenimage} & 12.494 & 0.3577 & 71.0277 & 0.0289 & 24.782 & 0.7954 & 25.9294 & 0.2246 & 22.037 & 0.7849 & 29.8410 & 0.2409 \\
Qwen-Image-Edit-2509~\cite{wu2025qwenimage} & 12.878 & 0.3955 & 67.1031 & 0.0264 & 21.926 & 0.6620 & 22.5778 & 0.1378 & 20.482 & 0.6548 & 25.8943 & 0.1506 \\
Qwen-Image-Edit-2511~\cite{wu2025qwenimage} & 14.971 & 0.4419 & 55.2527 & 0.0229 & 22.288 & 0.7776 & 20.4043 & 0.2140 & 21.585 & 0.7705 & 22.1430 & 0.2242 \\
Qwen-Image-Edit-2511~\cite{wu2025qwenimage} + LoRA & \topone{21.895} & \toptwo{0.5667} & \topone{24.0345} & \toptwo{0.0133} & \toptwo{39.108} & \toptwo{0.9736} & \toptwo{3.3396} & \toptwo{0.0241} & \toptwo{34.225} & \toptwo{0.9570} & \toptwo{5.6960} & \toptwo{0.0352} \\ \hline
\rowcolor{gray!10}\textbf{FreeDeshadow (Ours)} & \toptwo{21.553} & \topone{0.5692} & \toptwo{26.1266} & \topone{0.0107} & \topone{47.955} & \topone{0.9959} & \topone{1.8013} & \topone{0.0068} & \topone{37.766} & \topone{0.9847} & \topone{4.4735} & \topone{0.0156} \\
\hline
\end{tabular}}
\end{table*}

\subsection{Experimental Setup}

\subsubsection{Implementation Details}
We use Qwen-Image-Edit~\cite{wu2025qwenimage} (version $2511$), and fine-tune it with LoRA~\cite{hu2022lora} applied only to its attention projection layers.
Both the LoRA rank and the scaling factor $\alpha$ are set to $256$.
For optimization, we adopt the AdamW~\cite{loshchilov2019decoupled} optimizer with a learning rate of $1\times10^{-4}$, $\beta_1$ of $0.9$, $\beta_2$ of $0.999$, weight decay of $0.01$, and $\epsilon$ of $1\times10^{-6}$.
We use PyTorch's \texttt{ConstantLR} scheduler~\cite{paszke2019pytorch}: the learning rate is set to one-third of its base value for the first five optimization steps and then returns to $1\times10^{-4}$.
The model is trained in \texttt{bfloat16} mixed precision with a batch size of $8$ for $100{,}000$ steps.
Following the default settings of Qwen-Image-Edit, we train the model at a resolution of $1024 \times 1024$.
During inference, we use a $40$-step flow-matching sampler with the text guidance scale set to $4.0$.
All experiments are conducted on two NVIDIA A100 GPUs with 80 GB of memory each.

\subsubsection{Benchmarks}
To evaluate the adaptability of the model to diverse prompt cues in the shadow removal task, we constructed a new benchmark dataset for testing.
It covers six types of prompt cues: dot mask, line mask, object mask, object attribute textual, spatial relation textual, and object state textual.
The dataset contains $4,055$ valid cue-conditioned tasks in total, including $750$ dot prompts, $750$ line prompts, $750$ object mask prompts, $673$ object attribute prompts, $739$ spatial relation prompts, and $393$ object state prompts.
Dot, line, and object mask prompts test the ability to identify the target object from explicit spatial cues and associate it with its cast shadow.
Object attribute, spatial relation, and object state prompts test the ability to infer the shadow region of the target object from higher-level semantic descriptions.
In this way, the benchmark dataset provides a comprehensive assessment of the model across varying levels of cue abstraction.
In addition, all images in the dataset are provided at a resolution of $1024\times1024$, offering sufficient texture and structural detail for evaluating high-quality shadow removal.
To the best of our knowledge, this is the first benchmark to jointly support visual and textual prompts for controllable shadow removal, providing a solid evaluation foundation for future research.

\subsubsection{Baselines}
We compare FreeDeshadow with four categories of methods.
Specifically, 
\textbf{(i)} the first category consists of methods that take only visual prompts as input.
% accept only visual prompts. 
These methods perform shadow removal based on visual prompts such as an object mask. 
This category includes ControlNet-Inpainting~\cite{zhang2023controlnet}, FLUX.1 Fill [dev]~\cite{blackforestlabs2024fluxfilldev}, PowerPaint~\cite{zhuang2024powerpaint}, and HD-Painter~\cite{manukyan2025hdpainter}. 
\textbf{(ii)} The second category consists of methods that perform shadow removal solely from textual prompts.
This category includes SD1.5 Img2Img~\cite{stabilityai2022sd15}, InstructPix2Pix~\cite{brooks2023instructpix2pix}, SDXL-InstructPix2Pix~\cite{diffusers2023sdxlinstructpix2pix}, OmniGen~\cite{xiao2024omnigen}, UltraEdit~\cite{zhao2024ultraedit}, Step1X-Edit~\cite{liu2025step1xedit}, ICEdit~\cite{zhang2025icedit}, FluxEdit~\cite{paul2025fluxedit}, and FLUX.1-Kontext-dev~\cite{blackforestlabs2025fluxkontextdev}.
\textbf{(iii)} The third category consists of methods that support both visual and textual prompts, including FireRed-Image-Edit-1.1~\cite{fireredteam2025imageedit11}, FireRed-Image-Edit-1.0~\cite{fireredteam2025imageedit10}, FLUX.2 [dev]~\cite{blackforestlabs2025flux2dev}, ACE++~\cite{mao2025aceplusplus}, BrushEdit~\cite{li2026brushedit}, and different versions of Qwen-Image-Edit~\cite{wu2025qwenimage} (including Qwen-Image-Edit, Qwen-Image-Edit-2509, and Qwen-Image-Edit-2511).
For methods in this category, we evaluate all supported visual and textual prompt interfaces to comprehensively assess their shadow-removal performance across different prompt types. 
\textbf{(iv)} The fourth category consists of traditional shadow-removal methods that require a shadow mask as input, including
BMNet~\cite{zhu2022bijective}, Inpaint4Shadow~\cite{li2023leveraging}, ShadowFormer~\cite{guo2023shadowformer}, ShadowDiffusion~\cite{guo2023shadowdiffusion}, HomoFormer~\cite{xiao2024homoformer}, Token Painter~\cite{jiang2026token}, and SGCR-SR~\cite{beltrame2026winner}.
We also include our conference version PACSRNet~\cite{chen2025promptaware} for comparison.
All baselines are evaluated using their official implementations and released weights. 
These four categories together constitute comprehensive baselines, spanning traditional shadow removal methods as well as state-of-the-art visual-language image-editing models.

\subsubsection{Evaluation Metrics}
Following existing works~\cite{qu2017deshadownet,wang2018stcgan,le2019shadowdecomposition} on shadow removal, we adopt four widely used metrics to quantitatively evaluate the results of different models, including PSNR, SSIM, RMSE, and LPIPS.
Specifically, PSNR and SSIM measure the consistency between the removal result and the ground-truth image in terms of pixel fidelity and structural similarity. 
Higher PSNR and SSIM values ($\uparrow$) indicate that the result is closer to the ground-truth image.
RMSE measures the average pixel-level error between them, while LPIPS evaluates their perceptual difference at the deep-feature level.
Lower RMSE and LPIPS values ($\downarrow$) indicate smaller pixel-level error and smaller perceptual difference.
To evaluate the 
% restoration 
image quality in different regions, we further compute all the above metrics over three regions: the shadow region, the non-shadow region, and the full image. 
This scheme faithfully reflects the ability of a model to remove shadows for the user-specified objects as well as its capacity to preserve the original content in non-shadow regions.
Furthermore, we adopt IoU to evaluate the model's ability to accurately perceive the shadow cast by the user-specified object.
Since FreeDeshadow does not explicitly predict a shadow mask, we employ an auxiliary decoder to estimate the target-shadow mask from the object-aware shadow feature ${\textbf{\emph{G}}}$ generated by the cross-scale shadow grounding module.
The decoder consists of three blocks, each comprising a convolutional layer and an upsampling layer. 
Notably, the auxiliary decoder does not affect the shadow removal results and is used solely for evaluation and visualization. After FreeDeshadow has been trained, the auxiliary decoder is trained separately using BCE loss, with all parameters of FreeDeshadow frozen.
A higher IoU ($\uparrow$) indicates more accurate localization of the user-specified object's shadow.

\begin{figure*}[t!]
\centering
\includegraphics[width=1.0\linewidth]{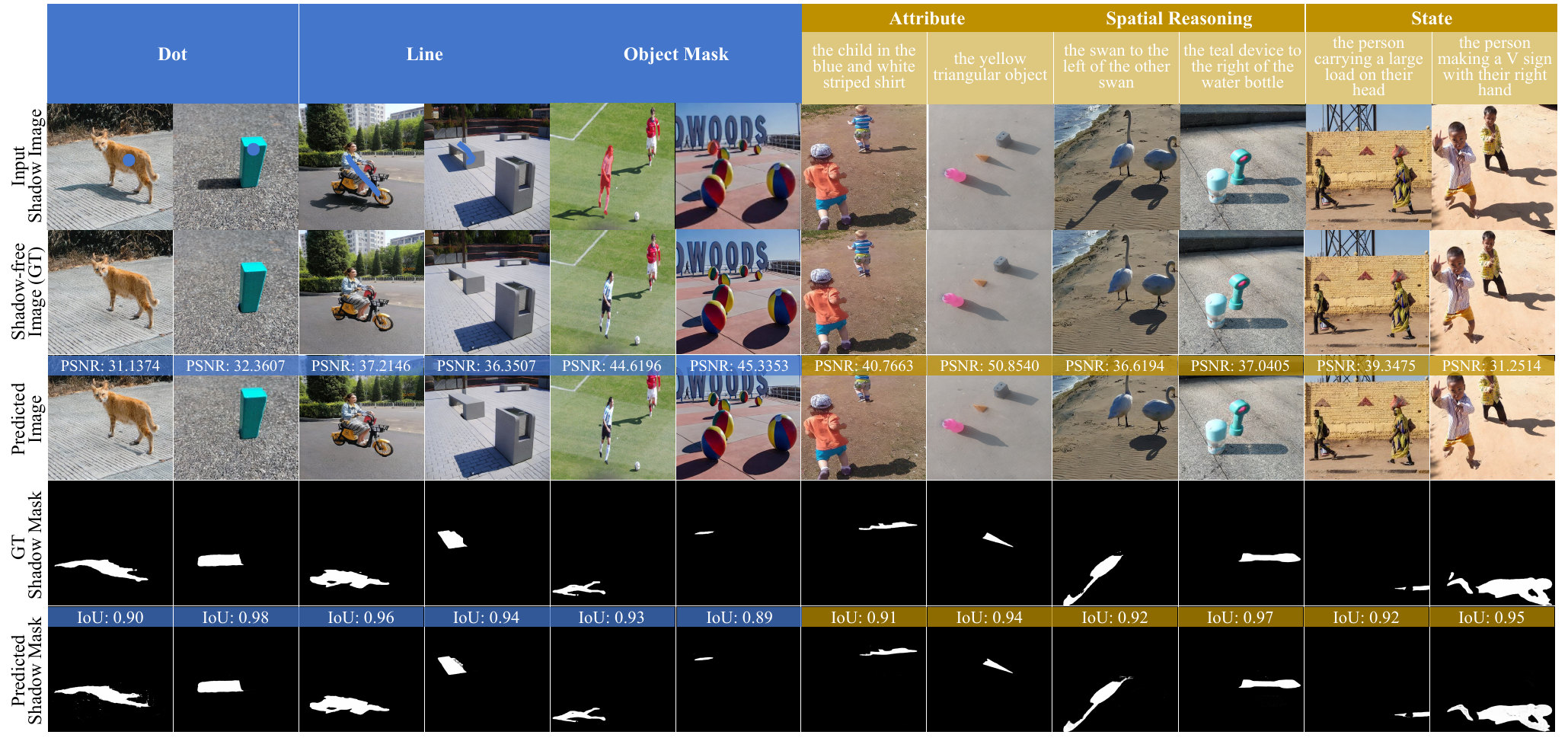}
\caption{
Qualitative results of FreeDeshadow under six visual and textual prompt types on the PCSRD+ dataset. 
The predicted target-shadow masks closely align with the corresponding ground-truth masks, demonstrating the model's ability to accurately localize the shadow cast by the user-specified object while preserving non-target content.
}
\label{fig:six_cue_qualitative}
\end{figure*}

\subsection{Experimental Results and Analysis}
\subsubsection{Quantitative Results}
Table~\ref{tab:main_model_metrics} reports the quantitative comparison on the PCSRD+ dataset under visual and textual prompts.
Under visual prompts, the proposed FreeDeshadow achieves the best results on nearly all metrics across the shadow, non-shadow, and all regions.
In the shadow region, PSNR rises from $15.069$ dB for the baseline FireRed-Image-Edit-1.0~\cite{fireredteam2025imageedit10} to $22.018$ dB, RMSE drops from $55.4391$ to $23.4455$, a reduction of about 58\%. 
Meanwhile, FreeDeshadow improves SSIM from the value of $0.4800$ achieved by PowerPaint~\cite{zhuang2024powerpaint} to $0.5756$.
In the non-shadow region, PSNR rises from $32.877$ dB for ACE++~\cite{mao2025aceplusplus} to $51.078$ dB, and LPIPS drops from $0.0387$ to $0.0037$, a reduction of more than $90$\%. 
Over all regions, PSNR improves over FireRed-Image-Edit-1.1~\cite{fireredteam2025imageedit11} by $13.875$ dB, and RMSE falls from $13.6644$ to $3.1272$, a reduction of about $77$\%. 
Notably, compared with the task-adapted Qwen-Image-Edit-2511 + LoRA baseline, FreeDeshadow further improves non-shadow and all region PSNR by $11.525$ dB and $5.864$ dB, respectively.
A similar advantage is observed under textual prompts.
FreeDeshadow achieves the best results on all non-shadow and all region metrics, while obtaining the best SSIM and LPIPS in the shadow region.
Compared with the baseline FLUX.1-Kontext-dev~\cite{blackforestlabs2025fluxkontextdev}, PSNR rises from $18.688$ dB to $21.553$ dB in the shadow region, and SSIM improves from $0.5139$ to $0.5692$. 
In the non-shadow region, PSNR gains $18.544$ dB, and LPIPS drops from $0.0482$ to $0.0068$, a reduction of about $86$\%. 
For the entire image, PSNR rises from $27.338$ dB to $37.766$ dB, and RMSE falls from $13.4358$ to $4.4735$, a reduction of about $67$\%. 
Even compared with Qwen-Image-Edit-2511 + LoRA, FreeDeshadow improves non-shadow and all region PSNR by $8.847$ dB and $3.541$ dB, respectively.
These results show that FreeDeshadow removes the shadow of the user-specified object under both visual and textual prompts.
It also preserves other content better than existing image editing models.

\begin{table}[t!]
\caption{Quantitative results on the PCSRD dataset~\cite{chen2025promptaware}. 
}
\vspace{-0.2cm}
\tabcolsep 8pt
\renewcommand\arraystretch{1.1}
\resizebox{1.0\linewidth}{!}{
\begin{tabular}{|r|cccc|}
\hline\thickhline
\rowcolor{gray!25}
Method & PSNR$\uparrow$ & SSIM$\uparrow$ & RMSE$\downarrow$ & LPIPS$\downarrow$ \\
\hline
BMNet~\cite{zhu2022bijective} & 22.224 & 0.7571 & 21.416 & 0.3517 \\
Inpaint4Shadow~\cite{li2023leveraging} & 27.602 & 0.7660 & 11.178 & 0.3434 \\
ShadowFormer~\cite{guo2023shadowformer} & 27.618 & 0.7668 & 11.172 & 0.3468 \\
ShadowDiffusion~\cite{guo2023shadowdiffusion} & 27.763 & 0.7698 & 11.060 & 0.3353 \\
HomoFormer~\cite{xiao2024homoformer} & 27.188 & 0.7633 & 11.594 & 0.3499 \\
PACSRNet~\cite{chen2025promptaware} & \toptwo{28.900} & 0.7898 & \toptwo{9.1525} & 0.2564 \\
Token Painter~\cite{jiang2026token} & 26.143 & \toptwo{0.9529} & 14.918 & \toptwo{0.0557} \\
SGCR-SR~\cite{beltrame2026winner} & 18.301 & 0.9141 & 31.392 & 0.1213 \\
\hline
\textbf{FreeDeshadow (Ours)} & \topone{39.383} & \topone{0.9863} & \topone{3.4490} & \topone{0.0131} \\
\hline
\end{tabular}}
\label{tab:results_pcsrd}
\end{table}

Furthermore, we report the quantitative comparison on the PCSRD dataset~\cite{chen2025promptaware} in Table~\ref{tab:results_pcsrd}.
As shown in the table, the proposed FreeDeshadow achieves the best results on PSNR, SSIM, RMSE, and LPIPS. 
In particular, FreeDeshadow achieves a PSNR of $39.383$ dB, outperforming our conference version PACSRNet~\cite{chen2025promptaware} by $10.483$ dB. 
Compared with Token Painter~\cite{jiang2026token}, SSIM improves from $0.9529$ to $0.9863$, while LPIPS drops from $0.0557$ to $0.0131$, a reduction of about $76$\%. 
Notably, FreeDeshadow requires only object-level visual prompts, such as dots, lines, or object masks, rather than an accurate shadow mask, yet still outperforms methods that rely on precise shadow masks as input. 
These results further demonstrate the effectiveness of FreeDeshadow for controllable shadow removal.

\begin{figure}[t!]
\centering
\includegraphics[width=1\linewidth]{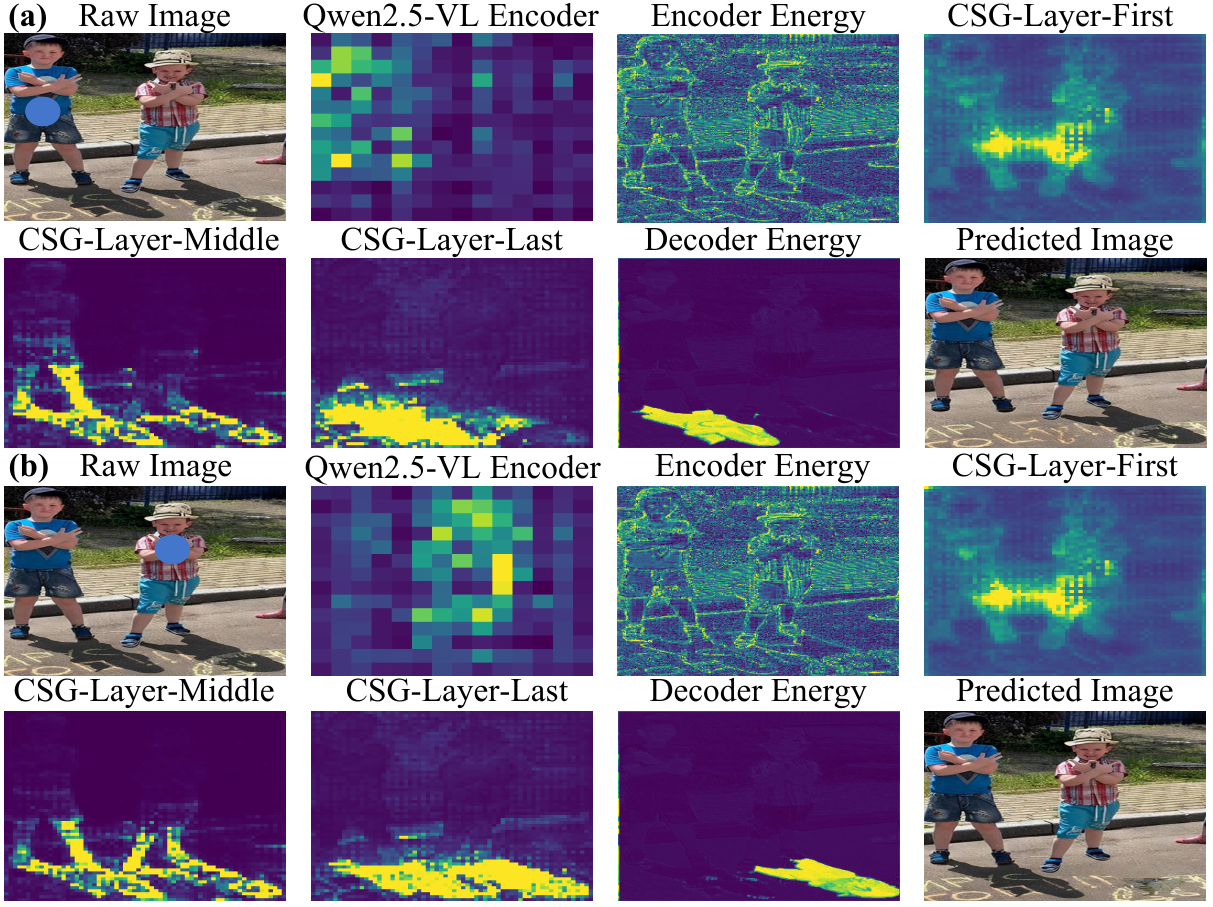}
\caption{Qualitative results of features in object-aware reasoning.
``CSG-Layer'' denotes the layer in the cross-scale shadow grounding module.}
\label{fig:vis_feature}
\end{figure}

\subsubsection{Qualitative Results}
Fig.~\ref{fig:six_cue_qualitative} presents the results of the proposed FreeDeshadow under three visual prompts (dot, line, object mask) and three textual prompts (attribute, spatial reasoning, state). 
For visual prompts, FreeDeshadow correctly associates the indicated object with its cast shadow and removes only its cast shadow, leaving the object, the shadows cast by other objects, and the surrounding background intact.
%leaving the object, the shadows cast by other objects, and the surrounding background.
For textual prompts involving attribute, spatial reasoning, or state, it parses the description and locates the corresponding shadow.
The resulting restorations generally achieve PSNR values above $30$ dB.
Furthermore, to evaluate the model's ability to identify the shadow cast by the user-specified object, we decode the object-aware shadow feature ${\textbf{\emph{G}}}$ generated by the cross-scale shadow grounding module using an auxiliary decoder. 
As shown in the last two rows of Fig.~\ref{fig:six_cue_qualitative}, the resulting masks closely align with the ground-truth masks, achieving an average IoU of $0.93$ over the twelve illustrated samples, even in scenes containing multiple objects and shadows.
These results demonstrate that FreeDeshadow can accurately interpret heterogeneous prompts and identify the correct shadow associated with the user-specified object.
To further investigate the shadow awareness of FreeDeshadow, Fig.~\ref{fig:vis_feature} visualizes the feature response in object-aware reasoning. 
As shown in the figure, the feature response is initially concentrated on the prompted object and progressively converges on its cast shadow, eventually aligning with the region restored by the model. 
Specifying the person on the left or right produces distinct response patterns that converge on their respective cast shadows.
Since the input image remains unchanged, these differences arise solely from the user prompts. 
This further demonstrates that FreeDeshadow can accurately distinguish the shadow cast by the user-specified object from other shadows, enabling precise object-level control and high-quality local restoration.

\begin{figure}[t!]
\centering
\includegraphics[width=1.0\linewidth]{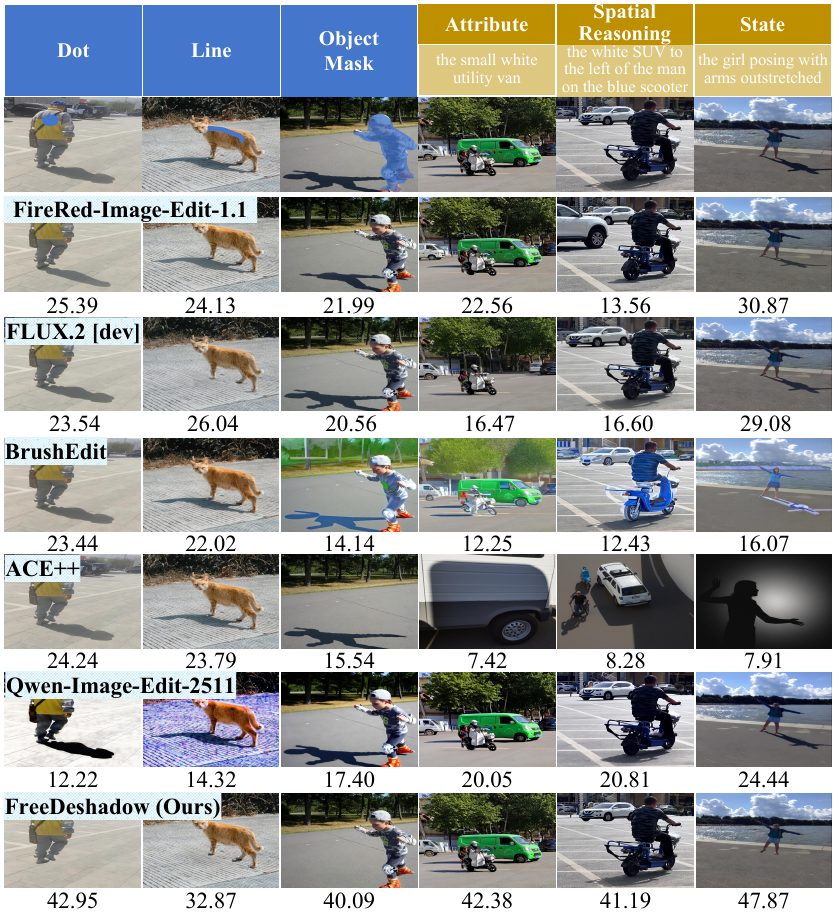}
\vspace{-0.6cm}
\caption{Qualitative comparison across six prompt types. 
The scalar denotes the all-region PSNR of each sample.
}
\label{fig:Qualitative comparison}
\vspace{-0.2cm}
\end{figure}

\begin{figure*}[t]
\centering
\includegraphics[width=1\linewidth]{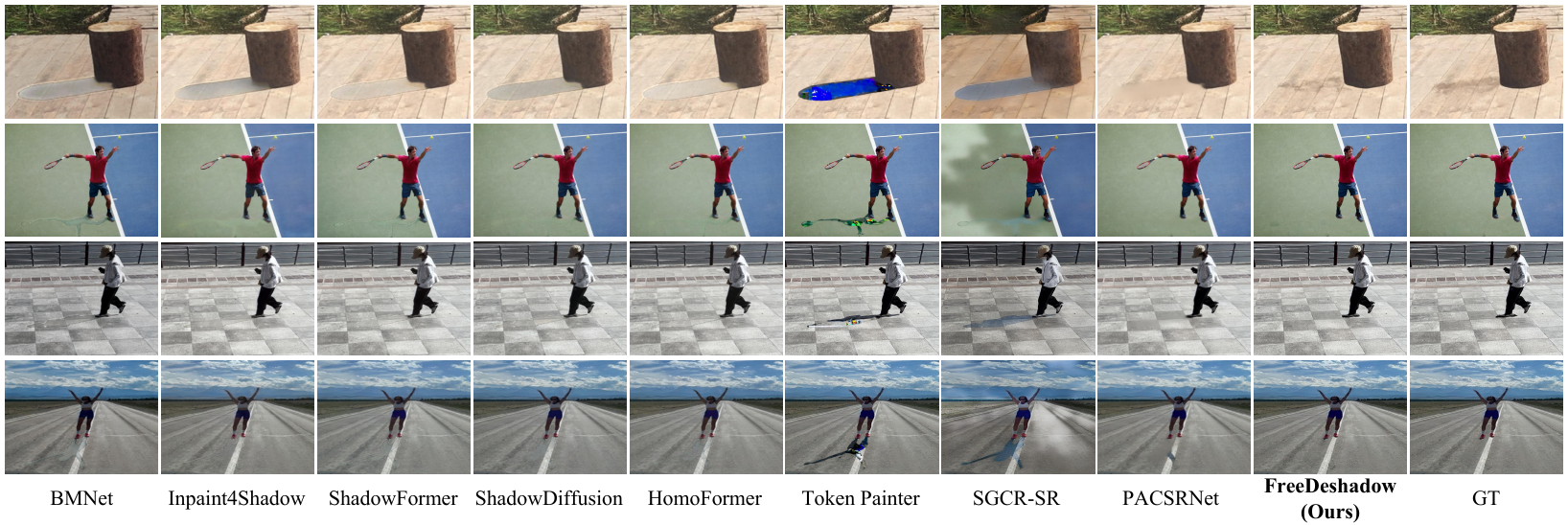}
\vspace{-0.8cm}
\caption{
Comparison with existing shadow removal methods.} 
\label{fig:vis_shadow_methods}
\end{figure*}

Fig.~\ref{fig:Qualitative comparison} compares FreeDeshadow with FireRed-Image-Edit-1.1~\cite{fireredteam2025imageedit11}, FLUX.2 [dev]~\cite{blackforestlabs2025flux2dev}, BrushEdit~\cite{li2026brushedit}, ACE++~\cite{mao2025aceplusplus}, and Qwen-Image-Edit-2511~\cite{wu2025qwenimage}. 
As shown in the figure, FireRed-Image-Edit-1.1~\cite{fireredteam2025imageedit11} and FLUX.2 [dev]~\cite{blackforestlabs2025flux2dev} often alter the color or shape of the object while removing the shadow. 
BrushEdit~\cite{li2026brushedit} and ACE++~\cite{mao2025aceplusplus} misjudge the shadow boundary, which damages large background areas or blacks out the image.
Qwen-Image-Edit-2511~\cite{wu2025qwenimage} introduces visible texture distortion and color shift. 
In contrast, the proposed FreeDeshadow accurately removes the shadows cast by the user-specified object under all six prompt types while preserving the object itself, other shadows, and the background.
These results demonstrate the advantage of FreeDeshadow in fine-grained and controllable shadow removal.
Furthermore, Fig.~\ref{fig:vis_shadow_methods} compares FreeDeshadow with existing shadow removal methods, including BMNet~\cite{zhu2022bijective}, Inpaint4Shadow~\cite{li2023leveraging},  ShadowFormer~\cite{guo2023shadowformer}, ShadowDiffusion~\cite{guo2023shadowdiffusion}, HomoFormer~\cite{xiao2024homoformer}, Token Painter~\cite{jiang2026token}, and SGCR-SR~\cite{beltrame2026winner}. 
Although FreeDeshadow uses only object-level visual prompts, such as dots, lines, and object masks rather than precise shadow masks, FreeDeshadow recovers finer texture details in the target shadow regions than methods that require precise shadow masks as input. 
This comparison demonstrates that FreeDeshadow achieves effective shadow removal without requiring explicit shadow-region specification.

\begin{figure}[t!]
\centering
\vspace{-0.2cm}
\includegraphics[width=1\linewidth]{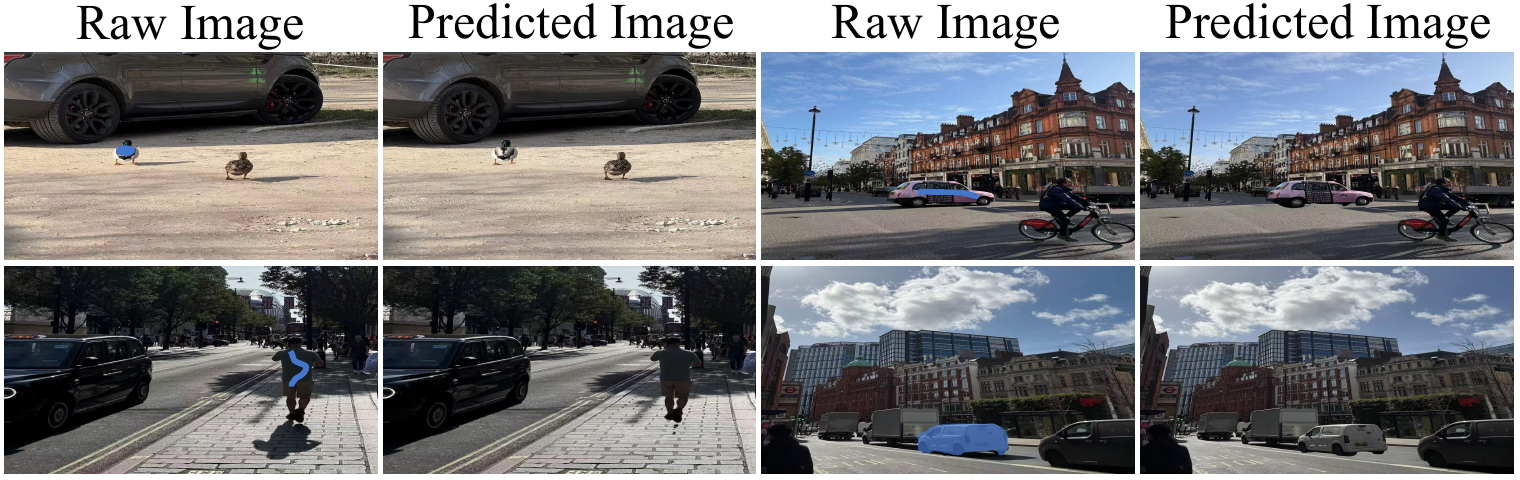}
\vspace{-0.6cm}
\caption{Real-image shadow removal under visual prompts.  
}
\label{fig:real_image_evaluation}
\vspace{-0.1cm}
\end{figure}

\subsubsection{Results on Real-World Scenarios}
To further evaluate the generalization of FreeDeshadow, we conduct experiments on real-world images. 
The test images cover diverse scenes, including parks, city streets, pedestrian walkways, and scenes with large sky regions. 
As shown in Fig.~\ref{fig:real_image_evaluation}, FreeDeshadow accurately localizes and removes target shadows of varying shapes and scales based on user prompts.
The restored shadow regions blend naturally with the surrounding illumination and colors, without visible boundary artifacts.
Furthermore, fine background details, such as building textures and pavement patterns, are well preserved, indicating that non-target image content remains largely unchanged.
These results demonstrate that FreeDeshadow generalizes well to real-world scenarios and remains robust across scenes and lighting conditions.

\subsubsection{User Study}
\begin{figure}[t!]
\centering
\vspace{-0.1cm}
\includegraphics[width=0.7\linewidth]{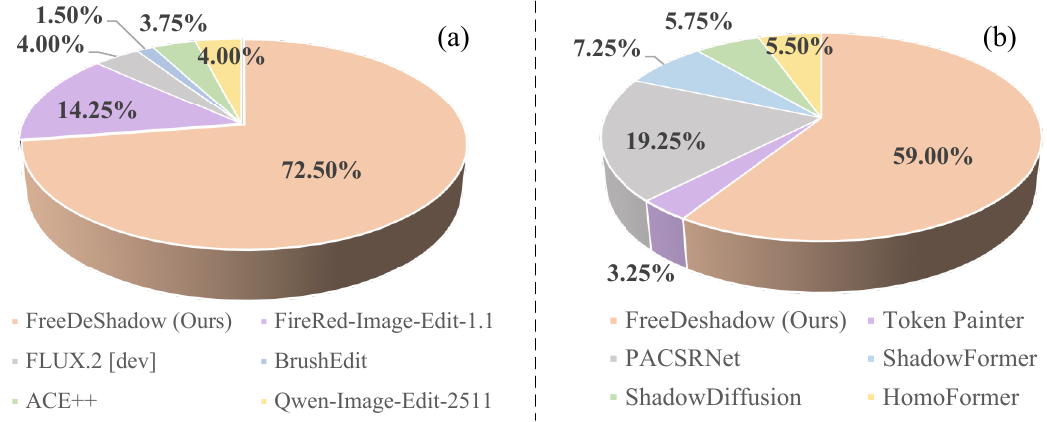}
\vspace{-0.2cm}
\caption{User Study. (a) Comparison with image editing methods. (b) Comparison with shadow removal methods. 
}
\label{fig:Fig_USS_2}
\end{figure}

To further verify the effectiveness of the proposed FreeDeshadow, we conduct a user study.
We first compare FreeDeshadow with five competitive image editing methods, including FireRed-Image-Edit-1.1~\cite{fireredteam2025imageedit11}, FLUX.2 [dev]~\cite{blackforestlabs2025flux2dev}, BrushEdit~\cite{li2026brushedit}, ACE++~\cite{mao2025aceplusplus}, and Qwen-Image-Edit-2511~\cite{wu2025qwenimage}. 
We select $10$ cases from the test dataset and invite $40$ volunteers to take part in the study.
For each case, we generate shadow removal results using all six methods and randomly shuffle them before presenting them to the participants, who then choose the result they consider the best.
As shown in Fig.~\ref{fig:Fig_USS_2} (a), FreeDeshadow receives $72.50$\% of the votes, substantially outperforming the five competing methods in user preference. 
These results indicate that FreeDeshadow outperforms general-purpose image editing models in faithfully removing the shadow cast by the user-specified object while preserving non-target content.

Furthermore, we compare FreeDeshadow with five shadow removal methods,
including HomoFormer~\cite{xiao2024homoformer}, ShadowDiffusion~\cite{guo2023shadowdiffusion}, Token Painter~\cite{jiang2026token}, ShadowFormer~\cite{guo2023shadowformer}, and PACSRNet~\cite{chen2025promptaware}. 
We follow the same evaluation protocol, resulting in $400$ votes. 
As shown in Fig.~\ref{fig:Fig_USS_2} (b), FreeDeshadow receives $59.00$\% of the votes, substantially outperforming the competing baselines in user preference. 
These results further demonstrate that FreeDeshadow is preferred by human observers over existing shadow removal methods.

\subsection{Analysis and Discussion}
In this section, we conduct ablation studies and provide additional analyses to evaluate the effectiveness of FreeDeshadow and to examine its practical applicability.
First, we ablate the individual modules of FreeDeshadow. 
By removing one module at a time, we analyze the contribution of each module to shadow removal performance. 
Then, we discuss the performance of FreeDeshadow in terms of multi-object editing, zero-shot cross-dataset generalization, controllability over the number of user-specified objects, and controllability over object selection, thereby demonstrating its applicability in more complex scenarios.

\subsubsection{Effectiveness of Shadow-Aware Local Refiner}
\begin{figure}[t!]
\centering
\includegraphics[width=1.0\linewidth]{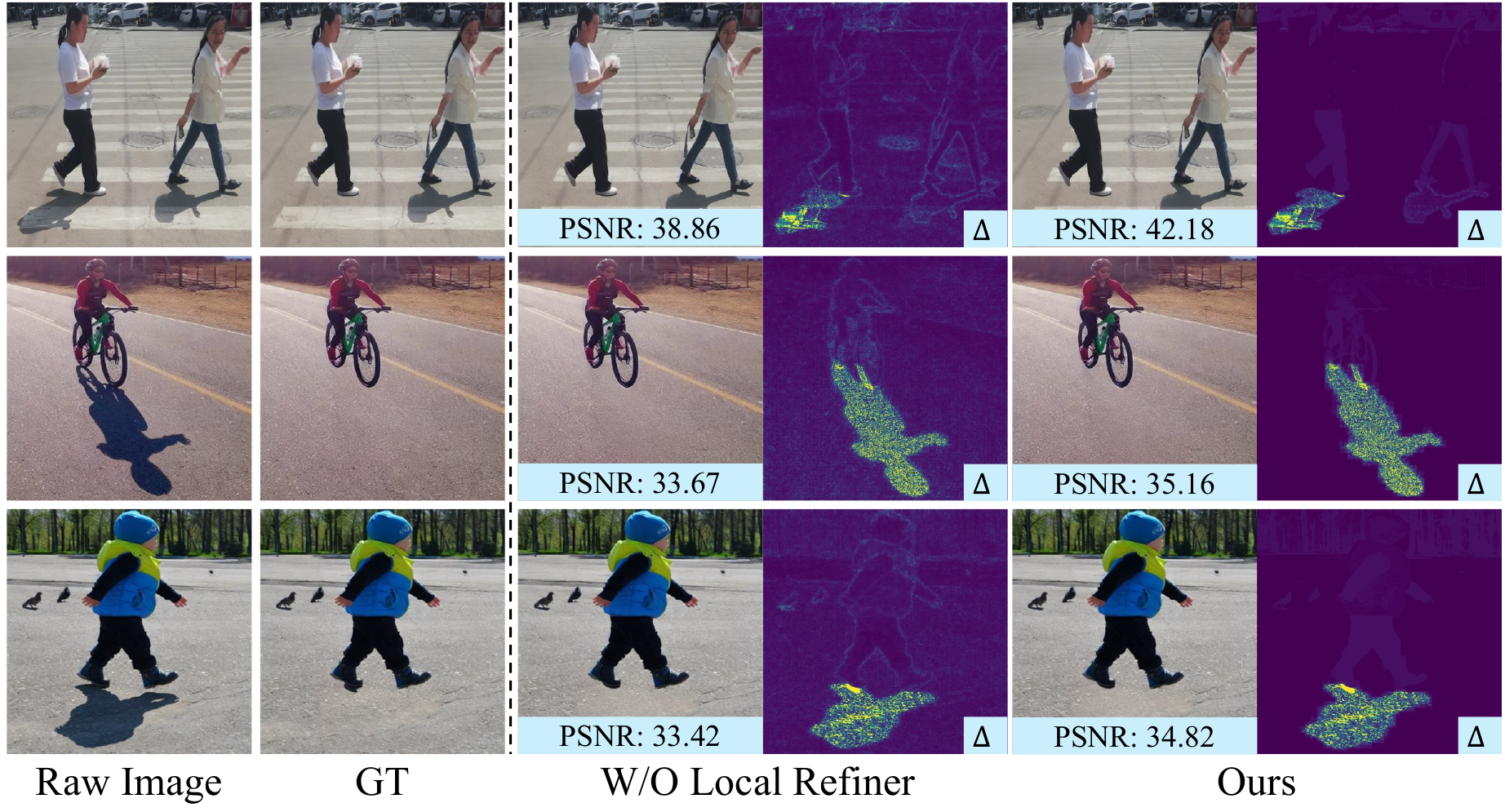}
\vspace{-0.6cm}
\caption{Effectiveness of the shadow-aware local refiner. ``$\Delta$'' denotes the difference between the result and ground-truth.}
\label{fig:Effectiveness_Refiner}
\end{figure}

To verify the effectiveness of the shadow-aware local refiner, we remove this module and decode the output of the object-aware generator directly, and compare it with the full model output. 
As shown in Fig.~\ref{fig:Effectiveness_Refiner}, decoding the output of the object-aware generator alone introduces color shifts and texture changes in the non-shadow regions, which are clearly visible in the difference maps. 
Adding the refiner keeps the surrounding content consistent with the input, recovers the shadow region more naturally, and yields higher PSNR. 
This is because the object-aware generator relies on a global diffusion process, which may introduce unintended changes outside the target shadow.
The refiner uses the object-aware shadow feature produced by the cross-scale shadow grounding module as spatial guidance.
It constrains the edits to the target shadow region, thereby restoring fine-grained local structures while preserving the surrounding content.
These demonstrate that the refiner is essential for ensuring localized and faithful editing.

\begin{table}[!t]
\centering
\caption{Module ablation on FreeDeshadow. 
\cmark{}/\xmark{} indicate that a component is enabled or ablated, respectively.}
\vspace{-0.2cm}
\label{tab:module_ablation}
\tabcolsep 4pt
\renewcommand\arraystretch{1.1}
\resizebox{0.99\linewidth}{!}{
\begin{tabular}{|cccc|ccc|}
\hline\thickhline
\rowcolor{gray!25} \makecell{Cue\\Cond.} & \makecell{Shadow\\Ground.} & \makecell{Prompt\\Adapter} & \makecell{Context\\Agg.} & \makecell{Shadow\\PSNR$\uparrow$} & \makecell{Non-Shadow\\PSNR$\uparrow$} &\makecell{Shadow\\IoU$\uparrow$}
\\
\hline\thickhline
\xmark & \cmark & \cmark & \cmark & 18.097 & 41.067 & 0.510 \\
\cmark & \xmark & \cmark & \cmark & 18.744 & \toptwo{49.496} & 0.051 \\
\cmark & \cmark & \xmark & \cmark & 20.221 & 49.208 & 0.121 \\
\cmark & \cmark & \cmark & \xmark & \toptwo{20.357} & 39.333 & \toptwo{0.819} \\
\cmark & \cmark & \cmark & \cmark & \topone{22.018} & \topone{51.078} & \topone{0.832} \\
\hline
%\vspace{-0.2cm}
\end{tabular}}
\vspace{-0.2cm}
\end{table}

\subsubsection{Effectiveness of Object-Aware Cue Conditioning}
Object-aware cue conditioning establishes an explicit correspondence between the user prompt and the specified object within the object-aware generator.
As shown in Table~\ref{tab:module_ablation}, removing this conditioning decreases the target-shadow localization IoU from $0.832$ to $0.510$, while reducing PSNR by about $3.9$ dB in the shadow region and $10.0$ dB in the non-shadow region. 
Without this conditioning, the model lacks reliable guidance for identifying the specified object and must infer the target shadow solely from the image content. 
This degrades both the restoration quality in the shadow region and the consistency of non-shadow regions. 
These results demonstrate the importance of object-aware cue conditioning for accurate target selection and controllable shadow removal.

\subsubsection{Effectiveness of Cross-Scale Shadow Grounding}
To evaluate the contribution of cross-scale shadow grounding, we remove this module and compare the resulting model with the full model.
As shown in Table~\ref{tab:module_ablation}, removing this module causes the target-shadow localization IoU to drop dramatically from $0.832$ to $0.051$, indicating a severe degradation in the model's ability to identify the shadow cast by the user-specified object. 
Meanwhile, the PSNR decreases by about $3.3$ dB in the shadow region and $1.6$ dB in the non-shadow region. 
Without cross-scale shadow grounding, the subsequent shadow-aware context aggregation lacks reliable target-region guidance, resulting in ineffective restoration of the target shadow. 
These results demonstrate that cross-scale shadow grounding plays a critical role in establishing accurate object–shadow correspondence.

\subsubsection{Effectiveness of Learnable Prompt Adapter}
The learnable prompt adapter maps the multimodal conditioning tokens describing the target object into a sparse prompt space and generates an object-aware prompt for subsequent cross-scale shadow grounding.
As shown in Table~\ref{tab:module_ablation}, removing this adapter decreases the target-shadow localization IoU from $0.832$ to $0.121$, while reducing the PSNR in non-shadow regions by about $1.9$ dB. 
These results demonstrate that the learnable prompt adapter effectively transfers object-specific semantic information to the cross-scale shadow grounding module, providing essential guidance for accurate target-shadow localization.

\begin{figure}[t!]
\centering
\includegraphics[width=1\linewidth]{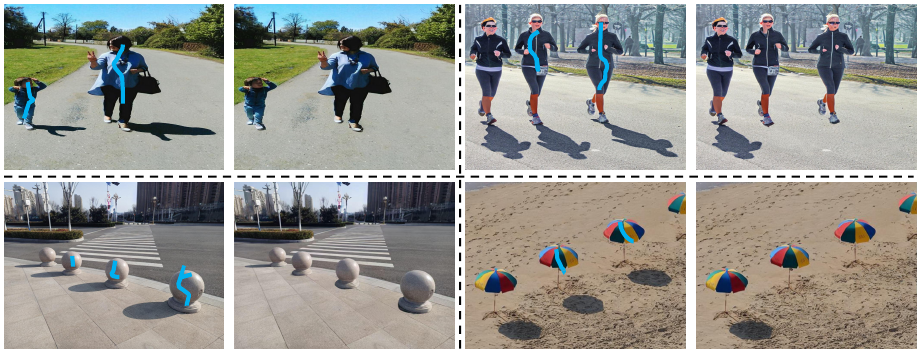}
\vspace{-0.6cm}
\caption{Visualization of multi-object shadow removal.}
\label{fig:multi_target_robustness}
\vspace{-0.3cm}
\end{figure}

\begin{figure}[t!]
\centering
\includegraphics[width=1.0\linewidth]{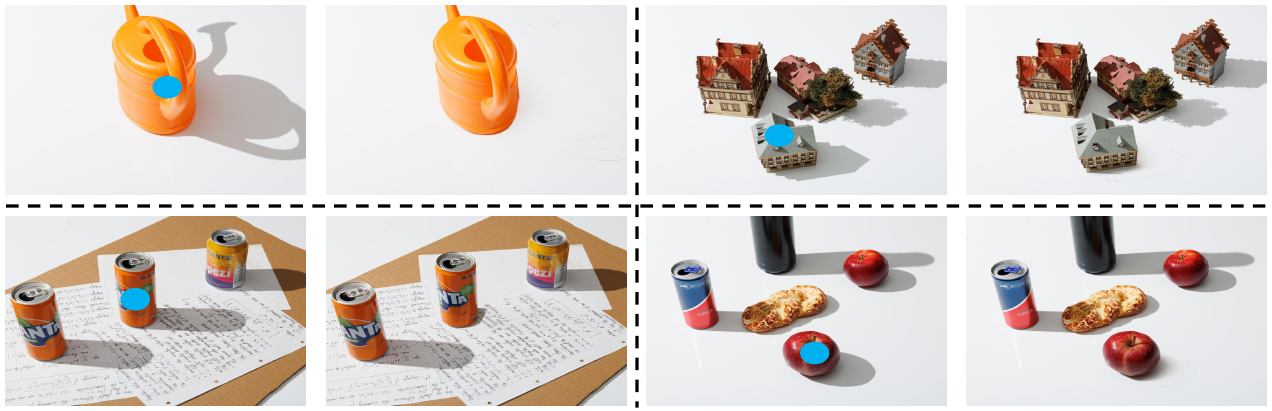}
\vspace{-0.6cm}
\caption{Zero-shot cross-dataset results. The model is evaluated on images from the WSRD dataset~\cite{Vasluianu_2023_CVPR} without fine-tuning.}
\label{fig:cross_dataset_zero_shot}
%\vspace{-0.1cm}
\end{figure}

\subsubsection{Effectiveness of Shadow-Aware Context Aggregation}
To evaluate the contribution of the shadow-aware context aggregation module, we remove it and feed the concatenation of the generator output and the object-aware shadow feature directly to the decoder.
As shown in Table~\ref{tab:module_ablation}, the PSNR in the non-shadow region drops by about $11.7$ dB, while the target-shadow localization IoU remains relatively stable ($0.819$ vs. $0.832$).
This indicates that the model can still localize the target shadow, but cannot effectively constrain the generated changes to the intended region. 
Without shadow-aware context aggregation, the decoder fails to suppress unintended modifications outside the target shadow.
These results demonstrate that this module is primarily responsible for spatially constrained restoration and non-target content preservation.

\subsubsection{Multi-Object Editing}
In practice, users often need to remove the shadows of multiple objects within the same scene.
Therefore, multi-object editing is an important capability for controllable shadow removal. 
Fig.~\ref{fig:multi_target_robustness} presents the multi-object editing results of FreeDeshadow.
The selected scenes contain diverse objects whose shadows are spatially adjacent or partially overlapping.
As shown in the figure, FreeDeshadow effectively removes the shadows of all specified objects, including shadow regions close to the objects and those extending farther away.
Meanwhile, the non-shadow regions such as road texture, grass and sand grains exhibit no visible color shift or structural distortion, indicating that the edits are effectively confined to the shadows associated with the specified objects.
These results demonstrate the robustness of FreeDeshadow in multi-object editing scenarios.
\begin{figure}[t!]
\centering
\includegraphics[width=1.0\linewidth]{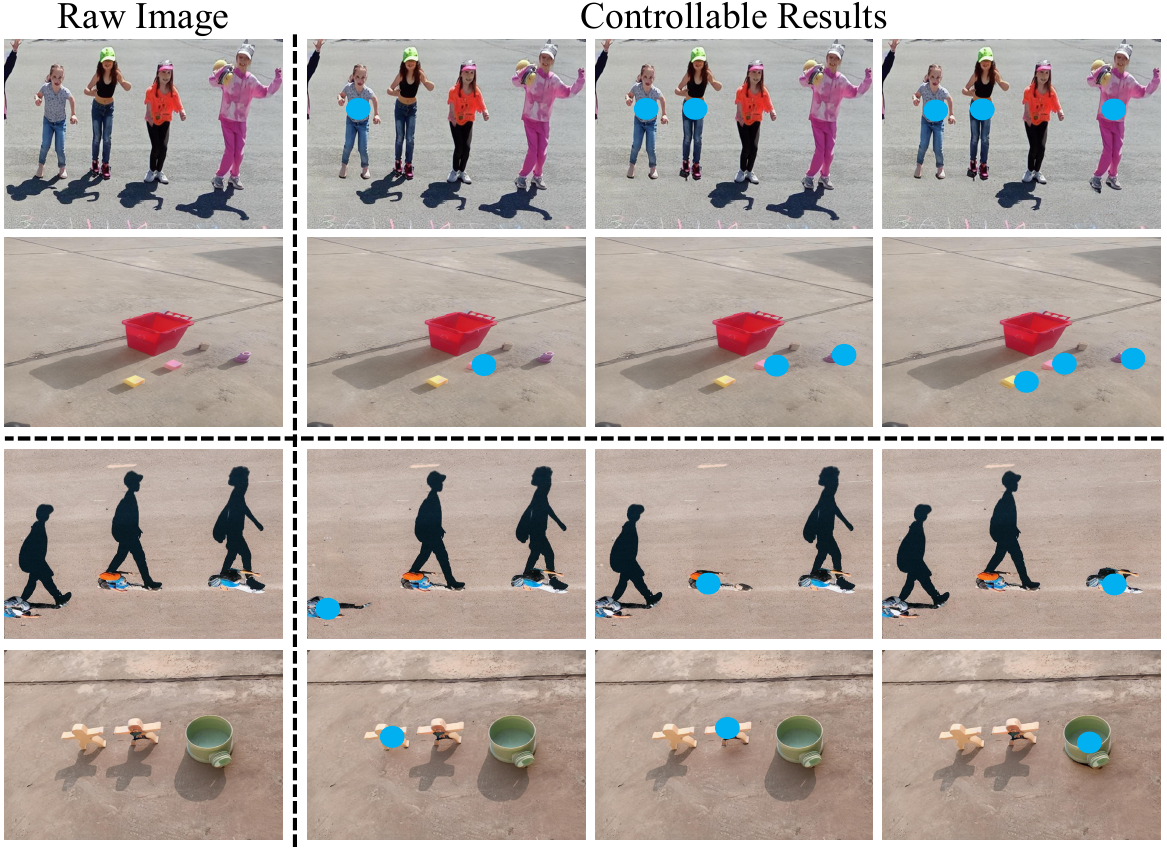}
\vspace{-0.6cm}
\caption{Qualitative results of FreeDeshadow’s controllability. }
\label{fig:controllability}
\end{figure}

\subsubsection{Zero-Shot Cross-Dataset Generalization}
To evaluate the cross-dataset generalization of FreeDeshadow, we apply the model trained on PCSRD+ to the WSRD dataset~\cite{Vasluianu_2023_CVPR} without any fine-tuning. 
As shown in Fig.~\ref{fig:cross_dataset_zero_shot}, FreeDeshadow can still identify the user-specified object and associate it with its cast shadow, effectively removing the corresponding shadow from unseen WSRD images.
In scenes containing multiple objects and cast shadows, FreeDeshadow consistently removes only the shadow cast by the user-specified object while preserving the other content.
This indicates that object-level controllability remains effective under domain shift.
Furthermore, on textured backgrounds such as paper with handwritten text, both the stroke structure and the paper texture within the shadowed region are well recovered by FreeDeshadow without noticeable blurring.
Across scenes with hard and soft shadow boundaries and varying background colors, the results show no noticeable color shift or residual shadow.
These observations suggest that the prompt-conditioned object–shadow association learned by FreeDeshadow is not limited to the data distribution of PCSRD+ and generalizes well across domains.

\subsubsection{Controllability Over the Number of User-Specified Objects}
The first two rows of Fig.~\ref{fig:controllability} show the shadow removal results of FreeDeshadow on the same scene as the number of user-specified objects gradually increases.
FreeDeshadow consistently removes only the shadows cast by the specified objects while preserving the remaining image content.
FreeDeshadow maintains stable shadow removal quality as more objects are selected.
These results demonstrate that FreeDeshadow supports flexible control over the number of objects to be edited, enabling fine-grained adjustment of the scope of shadow removal.

\subsubsection{Controllability Over Object Selection}
As shown in the last two rows of Fig.~\ref{fig:controllability}, we further evaluate FreeDeshadow's ability to distinguish the shadows cast by different objects in the same scene. 
By changing only the user-specified object while keeping the input image unchanged, FreeDeshadow produces different editing results that correspond to the selected objects' respective cast shadows. 
Meanwhile, the shadows of unselected objects and the surrounding background remain unchanged. 
These results further demonstrate that FreeDeshadow can establish object-specific shadow correspondence and provide precise object-level control.
\section{Conclusion}
\label{sec:conclusion}
This paper presents FreeDeshadow, a unified framework for object-aware controllable shadow removal. 
FreeDeshadow selectively removes the shadow cast by a user-specified object according to a visual or textual prompt, while preserving the object itself, other shadows, and the background content. 
To support training and evaluation, we further construct PCSRD+, a large-scale high-resolution dataset containing $154,571$ quadruples covering three visual and three textual prompts.
Extensive experiments demonstrate that FreeDeshadow achieves strong performance in both shadow removal quality and controllability. 
Beyond shadow removal, this work highlights a broader challenge in controllable image editing, where the user-specified object and the region to be edited are spatially separated and exhibit distinct visual characteristics. 
This object–effect discrepancy requires establishing accurate associations between objects and their corresponding visual effects while preserving unrelated scene content. 
Our study provides insights into this challenge and may inspire future research on other object-associated visual effects, such as reflections and localized illumination.

%\clearpage
\bibliographystyle{IEEEtran}
\bibliography{egbib}

\end{document}